\ifdefined\XeTeXversion\else\ifdefined\directlua\else\pdfoutput=1\fi\fi % arXiv: pdfLaTeX (keep in the first 5 lines)
\documentclass[11pt]{article}
\usepackage{thefinai}
\usepackage{latexsym,nicefrac}   % amssymb symbols come from newtxmath
\usepackage{url,colortbl,pifont,makecell,adjustbox,rotating}
\usepackage{listings}
\tcbuselibrary{listings,breakable,skins}
\usepackage{gradient-text}

\newtcblisting{promptlisting}[1]{
  enhanced,
  breakable,
  listing only,
  listing engine=listings,
  colback=finaicard,
  frame hidden,
  boxrule=0pt,
  arc=5pt,
  title={\small\sffamily\bfseries #1},
  coltitle=finaisoft,
  colbacktitle=finaicard,
  attach title to upper={\par\vspace{3pt}},
  listing options={
    basicstyle=\ttfamily\small,
    columns=fullflexible,
    keepspaces=true,
    showstringspaces=false,
    breaklines=true,
    breakatwhitespace=false,
    upquote=true,
    tabsize=2
  },
  left=8pt,
  right=8pt,
  top=6pt,
  bottom=6pt,
  before skip=6pt,
  after skip=6pt
}

\definecolor{bestblue}{RGB}{198,219,239}
\definecolor{secondblue}{RGB}{221,235,247}
\definecolor{bestred}{RGB}{252,187,161}
\definecolor{secondred}{RGB}{252,232,230}
\newcommand{\cmark}{\textcolor{green!60!black}{\ding{52}}}
\newcommand{\xmark}{\textcolor{red}{\ding{56}}}
\definecolor{lightblue}{RGB}{219,234,254}
\newcommand{\pmark}

\definecolor{upbest}{HTML}{C9D7F6}     % ~27% BLUE
\definecolor{upsecond}{HTML}{E6EEFC}   % overview light blue
\definecolor{downbest}{HTML}{DDCBF6}   % ~27% PURPLE
\definecolor{downsecond}{HTML}{F3ECFD} % overview light purple
\newcommand{\bestup}[1]{\cellcolor{upbest}\textbf{#1}}
\newcommand{\secondup}[1]{\cellcolor{upsecond}#1}
\newcommand{\bestdown}[1]{\cellcolor{downbest}\textbf{#1}}
\newcommand{\seconddown}[1]{\cellcolor{downsecond}#1}

\definecolor{tTrade}{HTML}{E0475B}
\definecolor{tHedge}{HTML}{8B3FD9}
\definecolor{tInsight}{HTML}{3567D8}
\definecolor{tAudit}{HTML}{1593B5}

\DeclareRobustCommand{\FAB}{\colorlet{fabouter}{.}\textsc{\gradientRGB{Herculean}{180,0,40}{120,0,240}}\color{fabouter}\xspace}

\title{\FAB: An Agentic Benchmark for Financial Intelligence}
\date{May 14, 2026 (first version); revised October 19, 2026}

\author{%
  Xueqing Peng\textsuperscript{1},
  Zhuohan Xie\textsuperscript{9},
  Yupeng Cao\textsuperscript{4},
  Haohang Li\textsuperscript{4},
  Lingfei Qian\textsuperscript{1},
  Yan Wang\textsuperscript{1},
  Vincent Jim Zhang\textsuperscript{1},
  Huan He\textsuperscript{1},
  Xuguang Ai\textsuperscript{1},
  Linhai Ma\textsuperscript{1},
  Ruoyu Xiang\textsuperscript{6},
  Yueru He\textsuperscript{3},
  Yi Han\textsuperscript{7},
  Shuyao Wang\textsuperscript{1},
  Yuqing Guo\textsuperscript{1},
  Mingyang Jiang\textsuperscript{1},
  Yilun Zhao\textsuperscript{2},
  Youzhong Dong\textsuperscript{1},
  Xiaoyu Wang\textsuperscript{6},
  Yankai Chen\textsuperscript{9,20},
  Ye Yuan\textsuperscript{20,21},
  Qiyuan Zhang\textsuperscript{9},
  Fuyuan Lyu\textsuperscript{20,21},
  Haolun Wu\textsuperscript{20,21},
  Yonghan Yang\textsuperscript{9},
  Zichen Zhao\textsuperscript{9},
  Yuyang Dai\textsuperscript{1},
  Fan Zhang\textsuperscript{9},
  Rania Elbadry\textsuperscript{9},
  Ayesha Gull\textsuperscript{1},
  Muhammad Usman Safder\textsuperscript{1},
  Nuo Chen\textsuperscript{16},
  Fengbin Zhu\textsuperscript{16},
  Tianshi Cai\textsuperscript{14},
  Zimu Wang\textsuperscript{14},
  Polydoros Giannouris\textsuperscript{18},
  Yuechen Jiang\textsuperscript{18},
  Zhiwei Liu\textsuperscript{18},
  Mohsinul Kabir\textsuperscript{18},
  Yuyan Wang\textsuperscript{18},
  Yixiang Zheng\textsuperscript{18},
  Yangyang Yu\textsuperscript{4},
  Weijin Liu\textsuperscript{4},
  Wenbo Cao\textsuperscript{1},
  Anke Xu\textsuperscript{1},
  Peng Lu\textsuperscript{10},
  Jerry Huang\textsuperscript{10},
  Mingquan Lin\textsuperscript{11},
  Prayag Tiwari\textsuperscript{17},
  Yijia Zhao\textsuperscript{12},
  V\'ictor Guti\'errez-Basulto\textsuperscript{19},
  Xiao-Yang Liu\textsuperscript{3},
  Kaleb E. Smith\textsuperscript{5},
  Jiahuan Pei\textsuperscript{15},
  Arman Cohan\textsuperscript{2},
  Jimin Huang\textsuperscript{1,10},
  Yuehua Tang\textsuperscript{8},
  Alejandro Lopez-Lira\textsuperscript{8},
  Xi Chen\textsuperscript{6},
  Xue Liu\textsuperscript{9,20,21},
  Junichi Tsujii\textsuperscript{13},
  Jian-Yun Nie\textsuperscript{10},
  Sophia Ananiadou\textsuperscript{18}%
}
\affiliation[1]{The Fin AI}
\affiliation[2]{Yale University}
\affiliation[3]{Columbia University}
\affiliation[4]{Stevens Institute of Technology}
\affiliation[5]{NVIDIA}
\affiliation[6]{New York University}
\affiliation[7]{Georgia Institute of Technology}
\affiliation[8]{University of Florida}
\affiliation[9]{MBZUAI}
\affiliation[10]{Universit\'e de Montr\'eal}
\affiliation[11]{University of Minnesota}
\affiliation[12]{University of Massachusetts Boston}
\affiliation[13]{National Institute of Advanced Industrial Science and Technology}
\affiliation[14]{University of Liverpool}
\affiliation[15]{Vrije Universiteit Amsterdam}
\affiliation[16]{National University of Singapore}
\affiliation[17]{Halmstad University}
\affiliation[18]{University of Manchester}
\affiliation[19]{Cardiff University}
\affiliation[20]{McGill University}
\affiliation[21]{Mila -- Quebec AI Institute}
\metadata[Code]{\ghlink{xueqingpeng/trading-analysis}}
\metadata[Data]{\hflink{datasets/TheFinAI/Herculean}}
\metadata[Leaderboard]{\href{https://xueqingpeng.github.io/herculean/}{xueqingpeng.github.io/herculean}}

\begin{document}

\begin{abstract}
As AI agents advance, the central question is no longer whether they
can solve isolated financial tasks, but whether their capabilities
generalize across complete professional workflows. Existing benchmarks
cannot answer this question because they evaluate either static
competencies or agents within a single interaction structure. We
introduce \FAB, a benchmark of end-to-end agentic financial intelligence
that crosses agent architecture, backbone model, and four structurally
distinct workflows: Trading, Hedging, Market Insights, and Auditing.
Each workflow is implemented as an MCP skill that preserves its native
state, actions, constraints, and outcome criteria, allowing intact agent
systems to be evaluated under matched external conditions. Across
\(30\) architecture--backbone configurations, we find that agent
capability is not globally ordered. Workflow rankings are weakly
aligned, changing the architecture frequently reverses performance
under the same backbone, and each architecture concentrates failure in
different forms of workflow control. These results establish that
agent performance is an interaction between architecture and workflow
demand, rather than an intrinsic property of the underlying model.
Consequently, success on a single financial workflow provides little
evidence of dependable, general financial intelligence.
% \footnote{Code: \url{https://github.com/xueqingpeng/trading-analysis}. \\ \hspace*{1.0em} Data: \url{https://huggingface.co/datasets/TheFinAI/Herculean}.}
\end{abstract}

\maketitle

\begin{figure}[t]
  \centering
  \includegraphics[width=\linewidth]{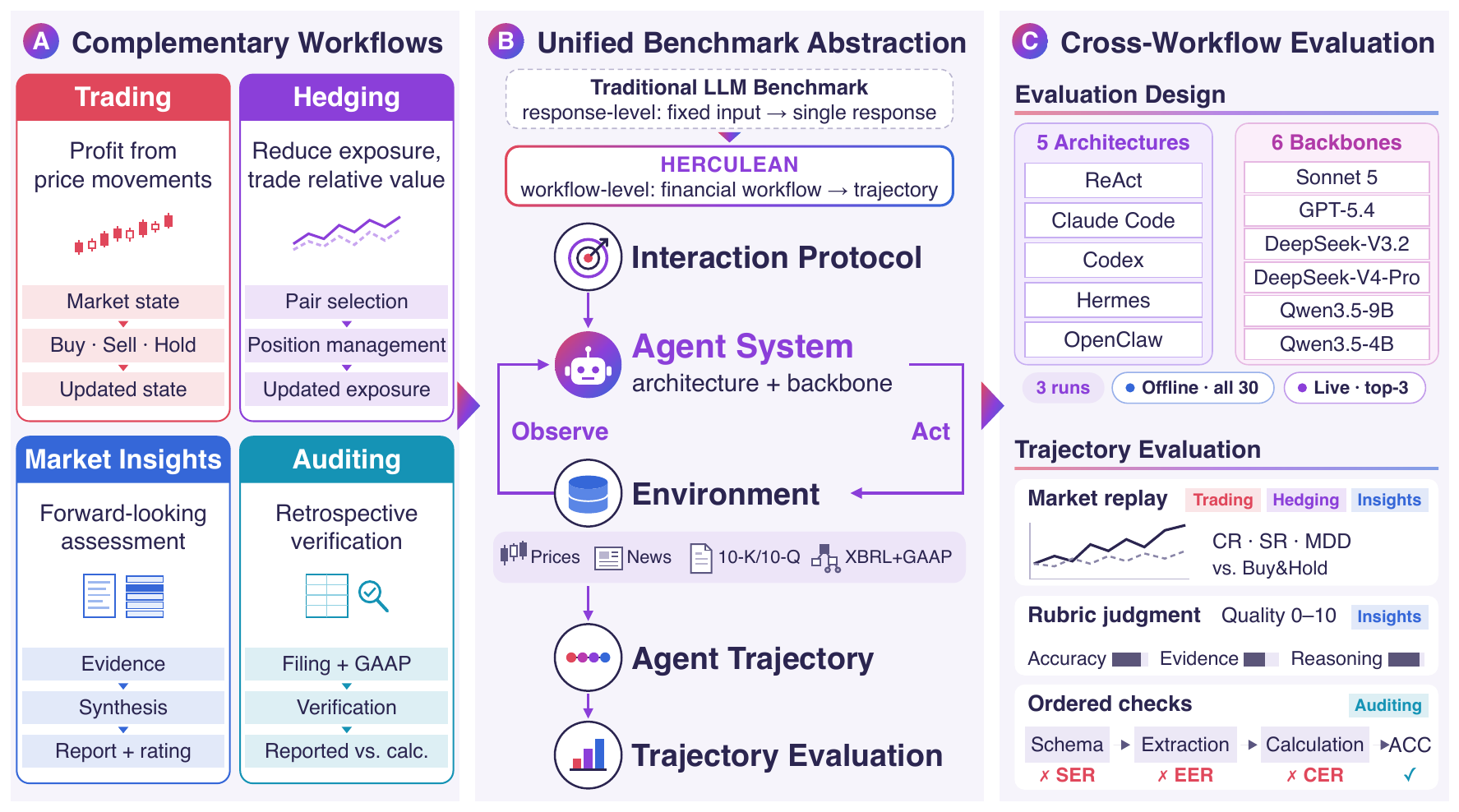}
  \caption{Overview of \FAB.}
  \label{fig:overview}
\end{figure}
\section{Introduction}
\label{sec:sec_introduction}

The next frontier for AI agents is not solving isolated tasks, but executing the complete workflows that define professional work. The central question is whether agents can operate end-to-end across professional workflows, ones where reasoning must continuously translate into action~\citep{yao2022react, schick2023toolformer}. Nowhere is this more consequential than in finance, where analysis has value only when it leads to commitment under uncertainty~\citep{lo2017adaptive}. Whether frontier agents can meet this standard is one of the most substantial tests of real-world agent intelligence.

However, whether current agents can operate reliably across
professional workflows with different operational demands remains
unclear. Financial LLM benchmarks evaluate static information-processing
tasks, including filings-based question answering, earnings-call
summarization, sentiment and risk classification, and document-level
evidence retrieval~\citep{
peng2025multifinbenbenchmarkinglargelanguage,
wang2026finauditingfinancialtaxonomystructuredmultidocument,
xie2023pixiu,NEURIPS2024_adb1d9fa}, but fixed inputs and outputs remove
the agent-controlled acquisition, action, and state management required
by complete workflows. Financial agent benchmarks introduce interactive
settings such as simulated trading, tool use, and financial
analysis~\citep{
li2025investorbench,qian2025agentstradelivemultimarket,
giannouris2026moiralanguagedrivenhierarchicalreinforcement,
fan2025ai,yu2025livetradebench}, but typically examine a single form of
financial work, leaving unclear whether an architecture's apparent
advantage persists across workflows with different execution demands.

To investigate this question, we introduce \FAB, a unified benchmark
for evaluating intact agents across four complete forms of professional
financial work: \textit{Trading}, \textit{Hedging},
\textit{Market Insights}, and \textit{Auditing}. Each workflow is
instantiated as a workflow-faithful skill environment that exposes its
own observations, tools, actions, state transitions, completion
conditions, and evaluation protocol through a common MCP interface.
The interface standardizes how agents access each workflow, while
leaving evidence acquisition, operation selection, state maintenance,
decision revision, and final commitment under agent control. Holding
the external conditions fixed within each workflow further enables
matched comparisons across agent architectures and backbone models.

The four workflows represent complementary dimensions of financial intelligence: profiting from market movements, managing market risk, forward-looking investment assessment, and retrospective financial verification.
\textit{Trading} requires long-horizon directional decisions over evolving
market states to profit from price movements~\citep{deng2016deeptrading,moody1998performance}.
\textit{Hedging} requires selecting related assets and maintaining
multi-asset positions as their relationship changes to reduce market exposure while profiting from relative price movements.
\textit{Market Insights} requires retrieving and synthesizing prices,
news, and filings into a structured investment report and committed
rating to inform forward-looking investment decisions~\citep{tetlock2007giving}.
\textit{Auditing} requires verifying SEC-style XBRL disclosures
~\citep{sec_xbrl} against the U.S. GAAP taxonomy through structured
retrieval and calculation-network reasoning to retrospectively identify accounting discrepancies.

Across five agent frameworks and six backbone models, \FAB reveals a
failure of capability transfer rather than a simple shortage of
capability. No system dominates all four workflows: each workflow has a
different leader, and rankings are only weakly aligned across workflows
(Kendall \(\tau\) between \(-0.23\) and \(0.38\)). Even the strongest
configuration, Hermes+DeepSeek-V4-Pro, reaches only \(53.5\) out of
\(100\) in its weakest workflow. With the backbone fixed, changing the
surrounding agent architecture can substantially alter both the
outcome and the dominant failure mode; in Auditing, for example,
DeepSeek-V3.2-based systems range from \(15.92\%\) to \(46.77\%\) accuracy,
and no backbone has a single best architecture across all four workflows.
Trajectory analysis further reveals distinct behavioral bottlenecks: agents fail through
qualitatively different pathways, including reactive decisions,
inconsistent state control, weak analysis-to-action transfer, and
invalid execution. As workflow demands change, architectural advantages
frequently disappear or reverse. Current agents therefore
exhibit workflow-contingent competence rather than a portable form of
end-to-end financial intelligence.

\section{Related Work}
% arXiv (single-column) layout of overleaf/tab_relatedwork.tex: same rows and caption, with the column headers of
% the per-benchmark table (Appendix A) written out in full. Keep the rows in sync with the ACL version.
\begin{table}[!b]
\centering
\footnotesize
\setlength{\tabcolsep}{5pt}
\begin{tabularx}{\linewidth}{p{5.2cm}|*{4}{>{\centering\arraybackslash}X}|*{2}{>{\centering\arraybackslash}X}c}
\toprule
& \multicolumn{4}{c|}{\textbf{Financial Workflow Coverage}}
& \multicolumn{3}{c}{\textbf{Evaluation Design}} \\
\textbf{Benchmark}
& \textbf{Trading}
& \textbf{Hedging}
& \textbf{\makecell{Market\\Insights}}
& \textbf{Auditing}
& \textbf{\makecell{Intact\\agent}}
& \textbf{\makecell{End-to-\\end}}
& \textbf{\makecell{Cross-\\workflow}} \\
\midrule
\raggedright AstaBench, General Agent Eval, FinDeepForecast, QFBench & \xmark & \xmark & \xmark & \xmark & \cmark & \xmark & \xmark \\
\raggedright FinBen, MultiFinBen & \cmark & \xmark & \xmark & \xmark & \xmark & \xmark & \xmark \\
\raggedright FinAuditing & \xmark & \xmark & \xmark & \cmark & \xmark & \xmark & \xmark \\
\raggedright FinMCP-Bench, Finance Agent Bench & \xmark & \xmark & \xmark & \xmark & \xmark & \xmark & \xmark \\
\raggedright FinRetrieval, FinDeepResearch, FINCH & \xmark & \xmark & \cmark & \xmark & \cmark & \xmark & \xmark \\
\raggedright InvestorBench, StockBench & \cmark & \xmark & \xmark & \xmark & \xmark & \cmark & \xmark \\
\raggedright Agent Market Arena & \cmark & \xmark & \xmark & \xmark & \cmark & \cmark & \xmark \\
\raggedright Moira & \xmark & \cmark & \xmark & \xmark & \xmark & \xmark & \xmark \\
\midrule
\rowcolor{lightblue}\textbf{\FAB\ (ours)} & \cmark & \cmark & \cmark & \cmark & \cmark & \cmark & \cmark \\
\bottomrule
\end{tabularx}
\caption{Workflow coverage and evaluation design of related benchmarks (rows group benchmarks with the same profile; per-benchmark details in Table~\ref{tab:benchmark_comparison_full}).}
\label{tab:benchmark_comparison}
\end{table}

Financial LLM benchmarks broaden task coverage behind fixed
model-facing interfaces~\citep{
DBLP:journals/corr/abs-2109-00122,
islam2023financebenchnewbenchmarkfinancial,NEURIPS2024_adb1d9fa,
zhang-etal-2025-xfinbench}; general agent benchmarks span interactive
domains without complete professional workflows~\citep{Wang2024GTA,mialon2024gaia,
bragg2025astabenchrigorousbenchmarkingai}; and financial agent benchmarks
add retrieval, tools, and trading environments~\citep{choi2025finagentbench,li2025investorbench,
qian2025agentstradelivemultimarket,bigeard2025fab,dong2025finch,
quantitativefinancebench2026} but do not evaluate how agent
architecture interacts with workflow structure.
Table~\ref{tab:benchmark_comparison} compares them with \FAB;
Appendix~\ref{sec:sec_relatedwork} gives the full discussion.

\begin{table}[b]
\centering
\footnotesize
\setlength{\tabcolsep}{2.7pt}
\begin{tabular*}{\linewidth}{@{\extracolsep{\fill}}l l l l c c l@{}}
\toprule
\textbf{Workflow} & \textbf{Evaluation Unit} & \textbf{Scope} & \textbf{History} & \makecell{\textbf{Offline}\\Jan--Mar} & \makecell{\textbf{Live}\\Aug--Sep} & \textbf{Metrics} \\
\midrule
Trading         & Asset trajectory     & 4 assets & from Dec 2024 & 244 (61$\times$4) & 168 (42$\times$4) & CR, SR, MDD \\
Hedging         & Universe trajectory & 8 assets, 28 pairs & from Dec 2024 & 61 (61$\times$1) & 42 (42$\times$1) & CR, SR, MDD \\
Market Insights & Asset trajectory     & 4 assets & from Dec 2024 & 52 (13$\times$4) & 36 (9$\times$4) & Quality; CR, SR, MDD \\
Auditing        & Filing--concept & 50 filings & FY2021--2024 & \multicolumn{2}{c}{67 cases (static)} & ACC, SER, EER, CER \\
\bottomrule
\end{tabular*}

\caption{
Evaluation scope: decisions per configuration and run (daily actions, weekly reports, or audit cases) in the offline and live windows of 2026; assets and data coverage are described in Section~\ref{sec:benchmark}.
}
\label{tab:eval_scope}
\end{table}

\section{\FAB Benchmark}
\label{sec:benchmark}

\FAB evaluates intact agent systems across four distinct yet
complementary financial workflows: Trading, Hedging, Market Insights,
and Auditing. Together, they span market profit-seeking and risk
management, forward-looking analysis and retrospective verification,
with sequential and bounded interactions.
\FAB provides a unified benchmark abstraction for workflow-level
evaluation, built on \emph{environment}, \emph{interaction
protocol}, and \emph{trajectory evaluation}.

\paragraph{Workflow-Level Evaluation}
The unit of evaluation in \FAB is a complete financial workflow rather than an
isolated model response. A workflow begins from an accessible financial
state, unfolds through agent-controlled interaction, and terminates in
an output whose validity is determined by the financial setting itself.
Evaluating only the final response overlooks the information acquisition,
execution, and completion conditions that determine whether an agent
can carry out the underlying work.

\paragraph{Unified Benchmark Abstraction}
We operationalize workflow-level evaluation through three coupled
elements: \emph{environment}, \emph{interaction protocol}, and
\emph{trajectory evaluation}. The \emph{environment} defines the
financial state available to the agent and the operations through which
it can be accessed. The \emph{interaction protocol} specifies the
objective and valid completion conditions, while leaving evidence
acquisition, operation selection, and stopping to the agent.
\emph{Trajectory evaluation} scores the resulting actions, artifacts,
or verification results using the criterion native to the workflow.
For workflow \(w\), instance \(j\), and agent \(A\), we write
\(\tau_{w,j}^{A}=\mathcal{R}_{w}(A,\mathcal{E}_{w,j})\) and
\(s_{w,j}^{A}=\mathcal{V}_{w}(\tau_{w,j}^{A})\),
where \(\mathcal{R}_{w}\) produces an agent--environment trajectory and
\(\mathcal{V}_{w}\) evaluates its committed outcome.

\paragraph{Shared Data Layer}
All four environments are built from a shared data layer across four modalities: daily OHLCV prices
with adjusted closes from Yahoo Finance~\citep{yfinance}; 10-K/10-Q filings from SEC
EDGAR~\citep{secapi}, from which the MD\&A and Risk Factors sections
are extracted; multi-source news~\citep{openai_websearch,finnhub,newsdata} consolidated into daily asset
records; and SEC XBRL filing packages, identified through XBRL US DQC
validation messages~\citep{xbrl_dqc},
paired with their referenced US-GAAP taxonomy releases. We normalize
tickers and metadata across sources, remove duplicate, amended, or
inaccessible records, and index every record by its public release
date, filings by their EDGAR \texttt{filed\_at} timestamp rather than
the fiscal period they report, preventing look-ahead leakage. Workflow-specific transformations then
convert this layer into executable environments without preselecting
the evidence or operations needed for completion
(Appendix~\ref{sec:sec_raw}). Table~\ref{tab:eval_scope} summarizes the evaluation scope of each workflow.

% \FAB realizes this
% paradigm across four substantially different financial workflows:
% Trading, Hedging, Market Insights, and Auditing. They span sequential
% and bounded interaction, actions and analytical artifacts, and
% outcome-based, judgment-based, and exact evaluation. We describe each
% workflow below through its \emph{environment}, \emph{interaction
% protocol}, and \emph{trajectory evaluation}.

\subsection{Trading}
\label{sec:sec_trading}

Trading tests whether an intact agent can sustain effective
decision-making as both an environment and the evidence available about
it evolve. Like a human trader, the agent must repeatedly determine
which newly available prices, news, and company disclosures are relevant
before committing to buy, sell, or hold.

\paragraph{Environment}
We construct a daily trading environment with offline and live
evaluation for four liquid U.S. equities: MSFT, TSLA, AAPL, and NVDA.
Offline evaluation replays a frozen historical collection from
December 1, 2024 to March 31, 2026, with approximately thirteen months
of queryable history followed by a three-month evaluation horizon.
Live evaluation runs from August 3 to September 30, 2026 and incorporates newly
released records before each daily interaction. Both modes use the same
assets, tools, and date-based access rules. The agent decides before
the market opens on each trading date and can access only records that
were public through the close of the previous trading date.

For each asset and date, the environment provides access to prices,
news, and corporate filings. Prices contain OHLCV records and adjusted
closes for actual U.S. trading days without imputation
(Appendix~\ref{sec:sec_price}); news is consolidated into
source-grounded daily asset records
(Appendix~\ref{sec:sec_news}); and 10-K and 10-Q filings provide their
Management's Discussion and Analysis and Risk Factors sections, with
coverage beginning in January 2024
(Appendix~\ref{sec:sec_filings}). The dedicated \texttt{trading\_mcp} server
exposes tools for historical prices and technical indicators, news
listing and retrieval, and filing listing and retrieval.

\paragraph{Interaction Protocol}
Each Trading instance pairs one asset \(i\) with an ordered sequence of
evaluation dates \(\mathbf{d}=(d_1,\ldots,d_T)\). The asset remains
fixed throughout the instance, isolating how the agent updates its
decisions over time from cross-asset selection and allocation. Given
the corresponding daily environments, the trajectory is
\(\tau_i^A=\mathcal{R}_{\mathrm{trade}}(A,\mathcal{E}_{i,\mathbf d})
=\bigl(\xi_{i,t}^A,a_{i,t}^A\bigr)_{t=1}^{T}\), where
\(a_{i,t}^A\in\{\textsc{Buy},\textsc{Sell},\textsc{Hold}\}\).

At each date \(d_t\), the \texttt{trading\_skill} implements the interaction protocol 
allowing the agent to commit exactly one action for asset
\(i\) before the environment advances to \(d_{t+1}\).
Full instructions for all four workflows are provided in
Appendix~\ref{sec:sec_prompts}.

\paragraph{Trajectory Evaluation}
Following prior work
~\citep{qian2025agentstradelivemultimarket}, we replay the committed
actions against the realized price path, with exposures \(+1\), \(-1\),
and \(0\) for \textsc{Buy}, \textsc{Sell}, and \textsc{Hold}, and
evaluate the strategy by cumulative return (CR), Sharpe ratio (SR), and
maximum drawdown (MDD), alongside a Buy\&Hold baseline
(Appendix~\ref{sec:sec_evaluation_metrics}).

\subsection{Hedging}
\label{sec:sec_hedging}

Hedging tests whether an intact agent can understand how a changing
environment jointly affects multiple entities and coordinate decisions
at two levels. Like a human hedger, the agent must first identify two
assets that tend to respond similarly to shared market conditions, and
then manage opposing positions as their behavior changes over time.
Because the selected pair defines the context for all later decisions,
pair selection and position management must be evaluated together.

\paragraph{Environment}
We construct a synchronized multi-asset environment over eight liquid
U.S. equities: AAPL, ADBE, AMZN, GOOGL, META, MSFT, NVDA, and TSLA,
yielding 28 candidate pairs. At each checkpoint, all assets are aligned
to the same trading date and information boundary, allowing candidate
relationships to be examined under a common market state.

The dedicated \texttt{hedging\_mcp} server exposes two complementary
query scopes. Asset-level tools support inspection of individual price
histories, technical indicators, news, and filings, while pair-level
tools support comparison of candidate assets over aligned market
records. The agent chooses which assets or pairs to inspect, the
historical windows and indicator parameters to use, and the textual
records to retrieve.

\paragraph{Interaction Protocol}
Under \texttt{hedging\_skill}, each Hedging instance consists of a candidate universe
\(\mathcal{U}\) and an ordered sequence of evaluation dates
\(\mathbf{d}=(d_1,\ldots,d_T)\). At the first checkpoint, the agent
selects an asset pair \(\pi^A=\{i^A,j^A\}\), which remains fixed
throughout the trajectory. It then makes one position decision for the
selected pair at each subsequent checkpoint, yielding
\(\tau^A=\bigl((\xi_1^A,\pi^A),
(\xi_t^A,a_t^A)_{t=2}^{T}\bigr)\), where
\(a_t^A\in\mathcal{A}_{\mathrm{hedge}}
=\{\textsc{Long--Short},\textsc{Short--Long},
\textsc{Hold},\textsc{Close}\}\).

For \(\pi^A=\{i^A,j^A\}\), \textsc{Long--Short} takes a long position
in \(i^A\) and a short position in \(j^A\), while
\textsc{Short--Long} reverses the two legs. \textsc{Hold} preserves the
current position, or remains flat if no position is open, and
\textsc{Close} exits the position. Open positions use equal absolute
notionals.

\paragraph{Trajectory Evaluation}
Following prior work
~\citep{giannouris2026moiralanguagedrivenhierarchicalreinforcement},
we replay the selected pair and subsequent position decisions as an
equal-notional long--short portfolio against the realized price paths
and report CR, SR, and MDD as in Trading.

\subsection{Market Insights}
\label{sec:sec_insights}

Market Insights tests whether an intact agent can turn heterogeneous
evidence into an explicit and internally consistent judgment. Like a
human analyst, the agent must determine which quantitative signals and
qualitative events matter, connect evidence across sources, and expose
the resulting analysis in a structured report rather than hide it
behind a single prediction, keeping the evidence, investment thesis,
risks, and final recommendation consistent.

\paragraph{Environment}
Using the shared market layer, we construct a weekly analytical
environment for MSFT, TSLA, AAPL, and NVDA. At each report week ending
on date \(m\), the environment includes prices, news, and corporate
filings available up to that date, together with a fixed peer mapping
for sector-relative comparison. A weekly aggregator computes 16
deterministic statistics covering asset performance, momentum, and
sector-relative behavior. These shared calculations provide a
comparable quantitative basis across agents without selecting the
qualitative evidence needed for the analysis.

The dedicated \texttt{market\_insights\_mcp} server exposes the weekly
aggregator, peer mapping, and query tools for prices, technical
indicators, news, and filings. Agents may inspect additional price
windows and indicator settings, choose the news horizon, and retrieve
selected filing sections.

\paragraph{Interaction Protocol}
Under \texttt{market\_insights} \texttt{\_skill}, each Market Insights instance pairs one asset \(i\) with an ordered
sequence of report weeks
\(\mathbf{m}=(m_1,\ldots,m_K)\). At each week \(m_k\), the agent
commits a structured report \(r_{i,k}^{A}\) together with an investment
rating \(g_{i,k}^{A}\). The complete trajectory is
\(\tau_i^A=(y_{i,k}^{A})_{k=1}^{K}\), where
\(y_{i,k}^{A}=(r_{i,k}^{A},g_{i,k}^{A})\) and
\(g_{i,k}^{A}\in\mathcal{G}
=\{\textsc{Strong Buy},\textsc{Buy},\textsc{Hold},
\textsc{Sell},\allowbreak \textsc{Strong Sell}\}\).

Each report follows a fixed eight-section structure covering the
executive summary; investment thesis and rating; price performance and
technical indicators; news and catalysts; earnings and filings;
sector-relative performance; risks; and outlook and scenarios. One
report is committed for each asset and report week and cannot be
revised after submission.

\paragraph{Trajectory Evaluation}
We evaluate two complementary aspects of each trajectory. First, the
weekly ratings are mapped to directional exposures and evaluated by CR,
SR, and MDD as in Trading.

Second, each report is evaluated using an LLM-generated,
instance-specific rubric covering three dimensions: content
accuracy, evidence fidelity, and reasoning quality. A judge LLM checks
the rubric criteria against the submitted report and its supporting
evidence. The three dimension scores are normalized to \(0\)--\(10\),
and the report-quality score for asset \(i\) is
\(Q_i^A=\frac{1}{3K}\sum_{k=1}^{K}\sum_{d=1}^{3}
q_d(r_{i,k}^{A})\); a missing report scores \(0\). 

We report the two aspects separately
(Appendix~\ref{sec:sec_evaluation_rubrics}).

\subsection{Auditing}
\label{sec:sec_auditing}

Auditing tests whether an intact agent can construct a reliable
verification procedure when the evidence is available but the path to
the answer is not given. Like a human auditor, the agent must identify
the relevant records, follow their relationships across documents, and
reconcile reported and calculated values, organizing multiple dependent
steps into a checkable procedure rather than retrieving a single fact.

\paragraph{Environment}
We construct a static verification environment containing 67 audit
instances from SEC XBRL filings paired with the corresponding US-GAAP
taxonomy. Following prior work
~\citep{wang2026finauditingfinancialtaxonomystructuredmultidocument},
we rebuild each instance from raw SEC EDGAR submissions rather than
provide pre-extracted evidence. Each filing is preserved as six linked
documents: the instance document, schema, calculation linkbase,
definition linkbase, label linkbase, and presentation linkbase. This
representation retains the filing contexts, dimensions, labels,
presentation structure, and calculation relations required for
verification.

The dedicated \texttt{auditing\_mcp} server exposes tools for resolving
the filing bundle, retrieving reported facts, traversing calculation
relations, and querying concept and taxonomy metadata. The environment
provides neither a prepared evidence chain nor a precomputed answer,
leaving the relevant relations and verification path to the agent.
Construction details are provided in
Appendix~\ref{sec:sec_xbrl} and
Appendix~\ref{sec:sec_gaap}.

\paragraph{Interaction Protocol}
Under \texttt{auditing\_skill}, each Auditing instance \(q\) specifies one filing, a target concept and
reporting period, and, where needed to identify a single fact, its
dimensional context or financial statement. Within a bounded interaction, the agent commits the
reported value \(\hat v_{q,\mathrm{rep}}^{A}\) and the value independently
reconstructed from the filing relationships
\(\hat v_{q,\mathrm{calc}}^{A}\), yielding
\(y_q^A=(\hat v_{q,\mathrm{rep}}^{A},
\hat v_{q,\mathrm{calc}}^{A})\) and
\(\tau_q^A=(\xi_q^A,y_q^A)\). 
The filing and target remain fixed throughout the instance, while the
retrieval and calculation procedure is left to the agent.

\paragraph{Trajectory Evaluation}
Following prior work
~\citep{wang2026finauditingfinancialtaxonomystructuredmultidocument},
we evaluate each committed value pair using deterministic, ordered
checks. An instance is classified as a schema error (SER) if the
submission is invalid, an extraction error (EER) if the reported value
is incorrect, a calculation error (CER) if the reconstructed value is
incorrect, and correct (ACC) only if all checks succeed. Because the
checks are sequential, the four rates sum to one
(Appendix~\ref{sec:sec_evaluation_metrics}).

\section{Experiments and Results}
\label{sec:sec_experiments}

\subsection{Evaluated Agents}
\label{sec:sec_agents}

\paragraph{Agents}
We select five
general-purpose frameworks that differ in how they plan, maintain
execution state, invoke tools, and recover from intermediate failures.
\textit{ReAct Agent} provides a minimal reasoning--action
loop~\citep{yao2022react}. \textit{Claude Code} and \textit{Codex}
represent execution-oriented agents with similar tool surfaces but
different planning and orchestration policies. \textit{Hermes}
represents a provider-agnostic autonomous loop, while
\textit{OpenClaw} uses a modular gateway that separates reasoning from
tool execution. This selection allows us to examine whether
framework-level execution control becomes more consequential as
workflows shift from bounded analysis to long-horizon decision-making
and deterministic verification. We exclude domain-specific financial
agents because their embedded task strategies would prevent the same
system design from being compared across workflows. Implementation
details are provided in Appendix~\ref{app:implement}.

% compact main-results table: mean (black) +- std (gray) over three runs; one headline metric per workflow.
% Source: herculean/eval (exports/final on HF: evaluation/). Per-asset metrics: tab_results_multirun_std.tex.
\begin{table}[t]
\centering
\footnotesize
\setlength{\tabcolsep}{0.6pt}
\renewcommand{\arraystretch}{1.25}
\begin{tabular*}{\linewidth}{@{\extracolsep{\fill}}l>{\centering\arraybackslash}p{38pt}>{\centering\arraybackslash}p{38pt}>{\centering\arraybackslash}p{38pt}>{\centering\arraybackslash}p{38pt}>{\centering\arraybackslash}p{38pt}@{\hspace{4pt}}c@{\hspace{4pt}}>{\centering\arraybackslash}p{38pt}>{\centering\arraybackslash}p{38pt}>{\centering\arraybackslash}p{38pt}>{\centering\arraybackslash}p{38pt}>{\centering\arraybackslash}p{38pt}@{}}
\toprule
 & \multicolumn{5}{c}{\textbf{\color{tTrade}Trading}} & & \multicolumn{5}{c}{\textbf{\color{tHedge}Hedging}} \\
\cmidrule(lr){2-6}\cmidrule(lr){8-12}
 & \multicolumn{1}{c}{\makecell{ReAct}} & \multicolumn{1}{c}{\makecell{Claude\\Code}} & \multicolumn{1}{c}{\makecell{Codex}} & \multicolumn{1}{c}{\makecell{Hermes}} & \multicolumn{1}{c}{\makecell{OpenClaw}} & & \multicolumn{1}{c}{\makecell{ReAct}} & \multicolumn{1}{c}{\makecell{Claude\\Code}} & \multicolumn{1}{c}{\makecell{Codex}} & \multicolumn{1}{c}{\makecell{Hermes}} & \multicolumn{1}{c}{\makecell{OpenClaw}} \\
\midrule
Sonnet 5 & $-$1.3{\color{gray}$\pm$.3} & $-$1.4{\color{gray}$\pm$.3} & $-$0.7{\color{gray}$\pm$.3} & $-$1.2{\color{gray}$\pm$.1} & $-$1.3{\color{gray}$\pm$.2} & & $-$1.3{\color{gray}$\pm$1.9} & $-$0.2{\color{gray}$\pm$2.9} & $-$2.6{\color{gray}$\pm$.8} & $-$0.5{\color{gray}$\pm$2.5} & $-$1.2{\color{gray}$\pm$2.1} \\
GPT-5.4 & $-$0.1{\color{gray}$\pm$.1} & \cellcolor{tTrade!12}0.4{\color{gray}$\pm$.9} & $-$1.2{\color{gray}$\pm$1.1} & $-$0.6{\color{gray}$\pm$.8} & \cellcolor{tTrade!30}\textbf{0.7}{\color{gray}$\pm$.1} & & 0.9{\color{gray}$\pm$1.8} & 1.0{\color{gray}$\pm$1.8} & $-$0.1{\color{gray}$\pm$1.9} & $-$1.2{\color{gray}$\pm$.9} & $-$0.4{\color{gray}$\pm$1.7} \\
DS-V3.2 & $-$0.2{\color{gray}$\pm$1.0} & $-$0.3{\color{gray}$\pm$.9} & $-$0.1{\color{gray}$\pm$.0} & $-$0.1{\color{gray}$\pm$.6} & $-$1.1{\color{gray}$\pm$1.3} & & $-$2.1{\color{gray}$\pm$.4} & $-$1.7{\color{gray}$\pm$.4} & $-$1.7{\color{gray}$\pm$1.5} & $-$1.7{\color{gray}$\pm$1.4} & $-$1.4{\color{gray}$\pm$1.2} \\
DS-V4-Pro & $-$0.7{\color{gray}$\pm$.2} & $-$0.4{\color{gray}$\pm$.4} & $-$1.5{\color{gray}$\pm$.4} & $-$0.3{\color{gray}$\pm$.4} & 0.3{\color{gray}$\pm$.8} & & $-$1.9{\color{gray}$\pm$.2} & $-$2.0{\color{gray}$\pm$.6} & $-$0.9{\color{gray}$\pm$1.0} & $-$0.2{\color{gray}$\pm$1.8} & $-$0.8{\color{gray}$\pm$1.4} \\
Qwen-9B & $-$1.2{\color{gray}$\pm$1.2} & $-$1.2{\color{gray}$\pm$.9} & $-$1.6{\color{gray}$\pm$.7} & $-$1.1{\color{gray}$\pm$.6} & $-$1.6{\color{gray}$\pm$.7} & & $-$1.8{\color{gray}$\pm$1.4} & \cellcolor{tHedge!30}\textbf{1.7}{\color{gray}$\pm$.4} & $-$1.2{\color{gray}$\pm$.9} & 0.8{\color{gray}$\pm$2.1} & $-$0.7{\color{gray}$\pm$.7} \\
Qwen-4B & $-$1.9{\color{gray}$\pm$.9} & $-$1.1{\color{gray}$\pm$.2} & $-$1.2{\color{gray}$\pm$.7} & $-$1.4{\color{gray}$\pm$.9} & $-$1.5{\color{gray}$\pm$1.0} & & $-$1.9{\color{gray}$\pm$1.6} & 1.1{\color{gray}$\pm$.6} & $-$0.5{\color{gray}$\pm$1.1} & 0.7{\color{gray}$\pm$1.8} & \cellcolor{tHedge!12}1.4{\color{gray}$\pm$1.0} \\
\midrule\addlinespace[2pt]
 & \multicolumn{5}{c}{\textbf{\color{tInsight}Market Insights}} & & \multicolumn{5}{c}{\textbf{\color{tAudit}Auditing}} \\
\cmidrule(lr){2-6}\cmidrule(lr){8-12}
 & \multicolumn{1}{c}{\makecell{ReAct}} & \multicolumn{1}{c}{\makecell{Claude\\Code}} & \multicolumn{1}{c}{\makecell{Codex}} & \multicolumn{1}{c}{\makecell{Hermes}} & \multicolumn{1}{c}{\makecell{OpenClaw}} & & \multicolumn{1}{c}{\makecell{ReAct}} & \multicolumn{1}{c}{\makecell{Claude\\Code}} & \multicolumn{1}{c}{\makecell{Codex}} & \multicolumn{1}{c}{\makecell{Hermes}} & \multicolumn{1}{c}{\makecell{OpenClaw}} \\
\midrule
Sonnet 5 & 8.2{\color{gray}$\pm$.1} & 8.4{\color{gray}$\pm$.1} & 8.4{\color{gray}$\pm$.1} & 8.4{\color{gray}$\pm$.1} & 8.5{\color{gray}$\pm$.0} & & 49.8{\color{gray}$\pm$2.3} & 44.3{\color{gray}$\pm$1.7} & 45.3{\color{gray}$\pm$.9} & 43.8{\color{gray}$\pm$3.8} & 48.8{\color{gray}$\pm$3.1} \\
GPT-5.4 & 8.5{\color{gray}$\pm$.0} & 8.6{\color{gray}$\pm$.1} & 8.4{\color{gray}$\pm$.1} & 8.5{\color{gray}$\pm$.1} & 8.4{\color{gray}$\pm$.0} & & 28.9{\color{gray}$\pm$3.1} & 26.4{\color{gray}$\pm$4.6} & 33.3{\color{gray}$\pm$3.1} & 32.8{\color{gray}$\pm$3.0} & 30.8{\color{gray}$\pm$1.7} \\
DS-V3.2 & 7.1{\color{gray}$\pm$.2} & 4.6{\color{gray}$\pm$.4} & 4.0{\color{gray}$\pm$.7} & 7.9{\color{gray}$\pm$.2} & 6.2{\color{gray}$\pm$.3} & & 15.9{\color{gray}$\pm$2.3} & 17.4{\color{gray}$\pm$1.7} & 42.3{\color{gray}$\pm$4.6} & 37.3{\color{gray}$\pm$3.0} & 46.8{\color{gray}$\pm$2.3} \\
DS-V4-Pro & 8.6{\color{gray}$\pm$.1} & \cellcolor{tInsight!12}8.7{\color{gray}$\pm$.1} & 8.4{\color{gray}$\pm$.0} & \cellcolor{tInsight!30}\textbf{8.8}{\color{gray}$\pm$.1} & \cellcolor{tInsight!30}\textbf{8.8}{\color{gray}$\pm$.0} & & 49.3{\color{gray}$\pm$3.0} & \cellcolor{tAudit!30}\textbf{54.7}{\color{gray}$\pm$3.8} & 50.7{\color{gray}$\pm$.0} & 51.7{\color{gray}$\pm$.9} & \cellcolor{tAudit!12}52.7{\color{gray}$\pm$.9} \\
Qwen-9B & 5.4{\color{gray}$\pm$2.9} & 6.7{\color{gray}$\pm$.9} & 7.4{\color{gray}$\pm$.1} & 7.0{\color{gray}$\pm$.4} & 6.4{\color{gray}$\pm$.2} & & 28.4{\color{gray}$\pm$12.9} & 13.9{\color{gray}$\pm$.9} & 28.4{\color{gray}$\pm$2.6} & 30.8{\color{gray}$\pm$.9} & 25.4{\color{gray}$\pm$.0} \\
Qwen-4B & 6.8{\color{gray}$\pm$.7} & 6.2{\color{gray}$\pm$.8} & 7.1{\color{gray}$\pm$.1} & 7.3{\color{gray}$\pm$.2} & 6.5{\color{gray}$\pm$.1} & & 28.9{\color{gray}$\pm$10.8} & 25.4{\color{gray}$\pm$.0} & 29.4{\color{gray}$\pm$2.3} & 29.4{\color{gray}$\pm$4.8} & 28.9{\color{gray}$\pm$2.3} \\
\bottomrule
\end{tabular*}
\caption{Main results (mean$\pm$std over three runs): Sharpe ratio for Trading and Hedging, report quality (0--10) for Market Insights, accuracy (\%) for Auditing. Best (bold) and second best are shaded. Full metrics in Appendix~\ref{app:multirun}.}
\label{tab:results}
\end{table}

\paragraph{Settings}
We pair every agent with six backbone models: two proprietary models,
Claude Sonnet 5
(\texttt{sonnet})~\citep{anthropic2026sonnet5} and GPT-5.4 (\texttt{gpt})~\citep{openai2026gpt54},
and four open-weight models, DeepSeek-V3.2 (\texttt{ds32})~\citep{deepseekai2025deepseekv32},
DeepSeek-V4-Pro (\texttt{ds4p})~\citep{deepseekai2026deepseekv4}, Qwen3.5-9B
(\texttt{qwen9})~\citep{qwen35_9b_hf}, and Qwen3.5-4B
(\texttt{qwen4})~\citep{qwen35_4b_hf}.
The complete \(5\times6\)
crossing yields \(30\) agent--backbone configurations per workflow.
Every configuration is run three times independently; we report the mean
and give standard deviations in Appendix~\ref{app:multirun}.
For the live window, which begins after every backbone was released
(Table~\ref{tab:agent_release_dates}) and thus mitigates parametric
look-ahead bias, we evaluate the three configurations per market
workflow with the highest mean offline Sharpe ratio, also three times
each (Appendix~\ref{sec:sec_live}).
Total inference
cost is approximately \$5{,}500~USD. Complete prompts and implementation
details are provided in Appendices~\ref{sec:sec_prompts}
and~\ref{app:implement}.

% All configurations face the same workflow settings with the corresponding \emph{environment}, \emph{interaction protocol}, and \emph{trajectory evaluation}, while retaining their
% native planning and tool-orchestration policies. Where supported,
% reasoning effort is set to \texttt{medium}. Persistent cross-run memory
% and built-in web search are disabled, preventing task information from
% entering outside the benchmark environment. Proprietary models are
% accessed through their official APIs and Qwen models through
% OpenRouter.\footnote{\url{https://openrouter.ai/}} Total inference
% cost is approximately \$4{,}000~USD. Complete prompts and implementation
% details are provided in Appendices~\ref{sec:sec_prompts}
% and~\ref{app:implement}.

\subsection{Main Findings}
\label{sec:sec_findings}

Table~\ref{tab:results} reports offline results for
\(30\) agent--backbone configurations across four financial
workflows, while Table~\ref{tab:live_std} (Appendix~\ref{sec:sec_live}) presents
live results.
%We organize the analysis around three questions:
% whether these workflows remain Herculean for current agents (RQ1),
% what behavioral signatures distinguish different agent architectures
% (RQ2), and how workflow environments expose architecture-specific
% strengths and failure modes (RQ3).
% , and whether these observations
% remain consistent across backbones and changing market conditions
% (RQ4).

\subsubsection{RQ1: Do End-to-End Financial Workflows Remain
Herculean for Current Agents?}
\label{sec:rq1}

\begin{figure}[t]
    \centering
    \includegraphics[width=0.891\linewidth]{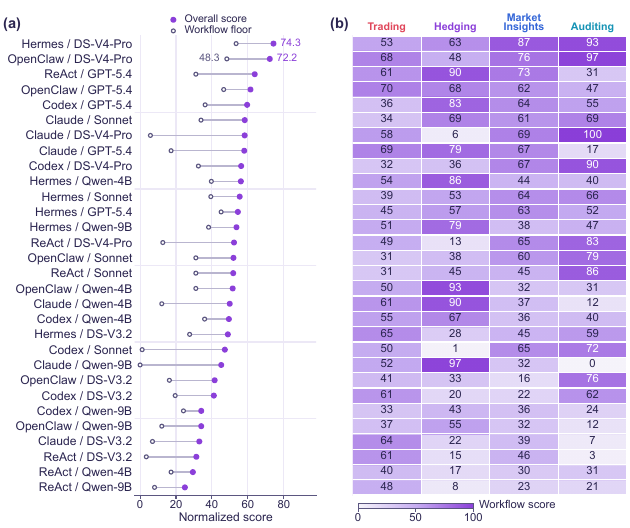}
    \caption{
    \emph{Fragmented capability} (\textbf{RQ1}): percentile-rank scores among the 30 configurations (0 = worst, 100 = best).
    \textbf{(a)} Overall score (mean of the four workflows) and workflow floor (lowest workflow).
    \textbf{(b)} Workflow scores (Appendix~\ref{app:rq1_rq3_scores}).
    }
    \label{fig:rq1_capability_landscape}
\end{figure}

No current system demonstrates reliable capability across the complete
benchmark. As shown in Figure~\ref{fig:rq1_capability_landscape},
Hermes+DeepSeek-V4-Pro achieves both the highest overall \FAB score
(\(74.3\)) and the strongest workflow floor, yet even this system
reaches only \(53.5\) in its weakest workflow, Trading. Moreover, different configurations lead the
four workflows, and their rankings are only weakly aligned across
workflows: Kendall \(\tau\) between workflow scores ranges from
\(-0.23\) to \(0.38\), and only Market Insights and Auditing are
significantly aligned (Appendix~\ref{app:rq1_rq3_scores}). The central limitation is thus not uniformly low capability,
but \emph{fragmented capability}: current agents reach strong
workflow-specific outcomes without forming a dominant and reliable
generalist.

The workflow-level results in Table~\ref{tab:results} reveal where this
fragmentation arises. In Trading, agents generally produce executable
trajectories, but no configuration achieves positive returns across all
four assets. Hedging is rarely profitable: only \(7\) of \(30\)
configurations obtain positive returns, and outcomes range
from \(-9.88\%\) to \(7.01\%\). Market Insights exhibits a different
gap: OpenClaw+DeepSeek-V4-Pro produces the strongest reports, averaging \(8.82\),
but its ratings yield a negative average return, showing that analytical
quality and decision quality need not coincide. Auditing remains
difficult even under deterministic evaluation: the highest accuracy is
only \(54.73\%\): most failures are calculation errors, and several
configurations also fail through invalid output structure.

\subsubsection{RQ2: How Does Agent Architecture Redistribute
Failure across Financial Workflows?}
\label{sec:rq2}

\begin{figure}[t]
    \centering
    \includegraphics[width=\linewidth]{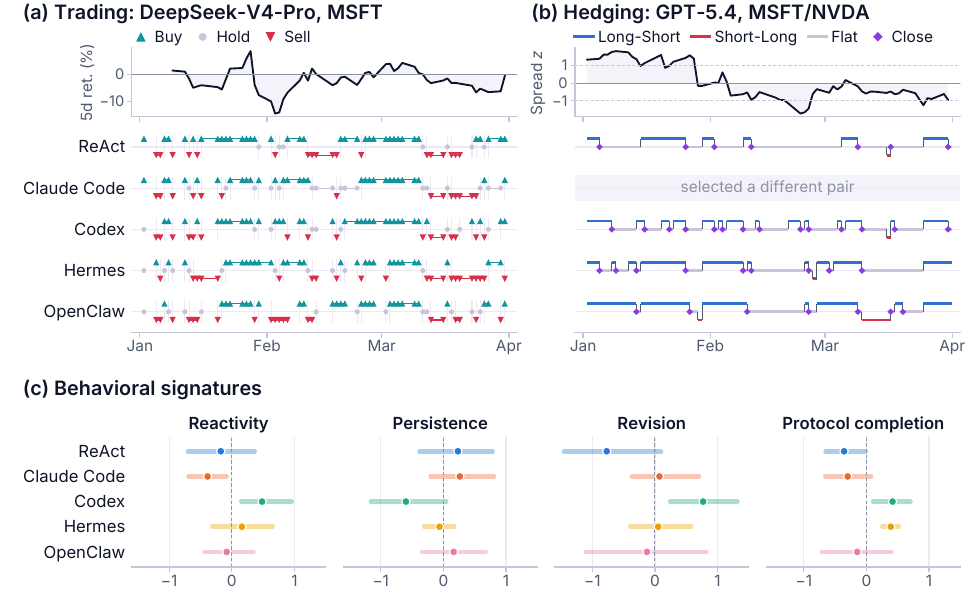}
    \caption{
    Architectures concentrate failure in different aspects of workflow control (\textbf{RQ2}).
    \textbf{(a)} Trading actions on MSFT under DeepSeek-V4-Pro (top: trailing 5-day return).
    \textbf{(b)} Hedging positions on MSFT/NVDA under GPT-5.4 (top: standardized spread).
    \textbf{(c)} Within-backbone standardized behavioral signatures with 95\% bootstrap intervals (Appendix~\ref{app:rq2_metrics}).
    }
    \label{fig:rq2_behavioral_traces}
\end{figure}

Outcome scores conceal qualitatively different failure pathways.
Across matched backbones, agent architectures redistribute observable
failures across four aspects of workflow control: evidence response (reactivity),
state maintenance (persistence), decision commitment (revision), and protocol execution (protocol completion).
Figure~\ref{fig:rq2_behavioral_traces} combines matched trajectories with
behavioral signatures to identify the dominant bottleneck associated with
each architecture.

Trading and Hedging expose distinct failures of sequential control.
Claude Code is the most conservative in Trading, assigning \(33.7\%\) of its
actions to \textsc{Hold}; Codex makes the most directional decisions (\(26.5\%\) \textsc{Hold}), while
ReAct switches most frequently (illustrated in Figure~\ref{fig:rq2_behavioral_traces}a). Yet every architecture relates more
strongly to past 5-day returns
(Spearman \(\rho=0.290\) to \(0.322\))
than to next-day returns
(\(\rho=-0.067\) to \(-0.040\)).
Their action policies differ, but all remain predominantly reactive.
In Hedging, profitable terminal outcomes can coexist with inconsistent
entry, position maintenance, reversal, or closure (Figure~\ref{fig:rq2_behavioral_traces}b), revealing failures
that aggregate return alone does not expose.

Market Insights and Auditing reveal different bottlenecks. Report
quality does not reliably predict the return of the committed rating
after controlling for asset, architecture, and backbone (OLS with fixed effects, $\beta=-0.249$, $p=0.19$), separating
analysis generation from decision commitment. In Auditing, structural
errors concentrate in ReAct (\(21.7\%\) on average, up to \(59.2\%\) with
DeepSeek-V3.2) and in Claude Code with DeepSeek-V3.2 (\(40.8\%\)), whereas
Codex and OpenClaw almost always preserve the required output structure
(\(3\%\) or less on average) and fail mainly at calculation.

Codex is the most reactive, revises its ratings most, and persists least; ReAct revises least; Claude Code is the least reactive; and Codex and Hermes complete the protocol most reliably (Figure~\ref{fig:rq2_behavioral_traces}c).
Evaluating only final
outcomes collapses these distinct pathways and obscures what an agent
architecture actually contributes.

\subsubsection{RQ3: How Do Workflow Demands Re-rank Agent
Architectures?}
\label{sec:rq3}

\begin{figure}[t]
    \centering
    \includegraphics[width=0.643\linewidth]{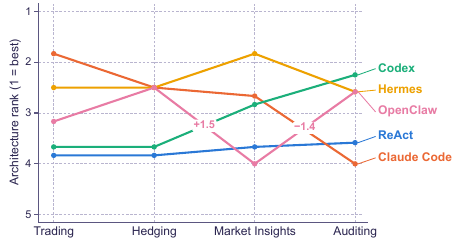}
    \caption{
    Workflow demands \emph{selectively re-rank} architectures (\textbf{RQ3}): mean within-backbone rank (1 = best; negative change = improvement; Appendix~\ref{app:rq1_rq3_scores}).
    }
    \label{fig:rq3_architecture_reranking}
\end{figure}

Architecture advantages are not portable across financial workflows.
Rather than uniformly increasing or decreasing difficulty, workflow
demands selectively re-rank agents according to the form of control
they require. As shown in Figure~\ref{fig:rq3_architecture_reranking},
architectures that perform strongly under one demand frequently lose
their advantage when the environment shifts from temporal and stateful
control to commitment and protocol control.
For example, Claude Code ranks first in Trading (mean rank 1.83)
but last in Auditing (4.00), whereas Codex ranks
near the bottom in Trading and Hedging (3.67 in both) but first in
Auditing (2.25). Within a backbone, the architecture ordering in one
workflow is negatively correlated with that in another for \(16\) of the
\(36\) backbone--workflow pairs, and the best architecture changes
across workflows under every backbone.

Trading and Hedging require agents to repeatedly revise decisions while
maintaining exposure over time, yet they do not favor the same systems.
For example, Claude Code+DeepSeek-V4-Pro attains a Trading score of \(57.8\)
but only \(5.8\) in Hedging, whereas Codex+GPT-5.4 reaches \(82.8\) in
Hedging but only \(36.2\) in Trading (Figure~\ref{fig:rq1_capability_landscape}). Adding persistent multi-entity
state therefore changes which execution policy is effective, rather
than simply making the same architectures uniformly weaker.

The re-ranking becomes sharper under commitment and protocol demands.
Codex+DeepSeek-V4-Pro achieves \(89.7\) in Auditing but only \(32.5\) in
Trading, while Claude Code+Qwen3.5-9B falls from \(96.6\) in Hedging to \(0.0\)
in Auditing (Figure~\ref{fig:rq1_capability_landscape}). 
Market Insights introduces another ordering: systems that
generate strong reports do not necessarily produce effective committed
ratings. Advantages in sequential action control therefore do not
reliably transfer to analysis-to-action consistency or constrained
execution.

Workflow environments thus act as \emph{selective filters} over agent
architectures. Temporal adaptation, persistent state, consequential
commitment, and protocol compliance reward different execution
properties, producing weak alignment between workflow rankings.
Architecture quality is therefore conditional on the control demand
being tested, and conclusions drawn from a single financial workflow
do not generalize to end-to-end agent capability.

\section{Conclusion}

We introduced \FAB, a benchmark for evaluating intact financial agents
across four structurally distinct professional workflows. By crossing
agent architecture, backbone model, and workflow under matched
conditions, \FAB reveals that financial agent capability is not globally
ordered. No system dominates across workflows, architectural advantages
frequently disappear or reverse, and different architectures concentrate
failure in different aspects of workflow control. These findings show
that success in a single setting provides limited evidence of dependable
financial intelligence. Reliable financial agents will therefore require
architectures that can sustain decision making, state, commitment, and
execution across heterogeneous end-to-end workflows.

\section*{Limitations}
\label{sec:sec_limitations}

\FAB has several limitations.
First, the four workflows omit other important financial tasks, such as credit risk assessment, compliance, portfolio optimization, and fraud detection.
Second, the evaluation covers large-cap U.S.\ equities, English-language disclosures, and U.S.\ GAAP over fixed windows of a few months, so it may not reflect other markets, instruments, or economic cycles.
Third, the environments omit macroeconomic signals and frictions such as transaction costs and latency, which may overstate agent effectiveness.
Fourth, date-bounded access prevents look-ahead within the environment, but backtests cannot rule out information memorized by the backbones; the post-release live evaluation mitigates this risk over a shorter window.
Fifth, inference cost limits the number of assets, agent architectures, and backbone models.
Finally, report quality is scored by an LLM judge against LLM-generated rubrics that are not manually validated; although a second judge agrees closely (Appendix~\ref{sec:sec_evaluation_rubrics}), both may favor surface fluency over subtle reasoning errors.

\section*{Ethical Concerns}
\label{sec:sec_ethicalconcerns}

\paragraph{Data and intended use} The authors take full responsibility for the development and dissemination of \FAB and all related materials. All data are drawn from publicly available sources (market prices, SEC filings, and summarized news) and contain no personal or sensitive information; the design, construction, and public release of \FAB adhere to established ethical standards, applicable data licensing terms, and privacy requirements. \FAB is intended strictly for academic research, model evaluation, and methodological development. Neither the benchmark nor any associated source code, datasets, or supplementary materials should be interpreted as financial, legal, or investment advice; users are strongly encouraged to consult qualified professionals before making financial or investment decisions, and the authors disclaim responsibility for any losses, damages, or other consequences arising from the use of, or reliance on, \FAB in practical or commercial settings.
\paragraph{Positive impacts} \FAB offers a rigorous, workflow-level evaluation protocol that complements static QA-style benchmarks, surfacing failure modes such as deterministic-calculation errors in \textit{Auditing} and weak signal-to-action translation in \textit{Market Insights}. By doing so, it guides research toward more trustworthy financial agents and, through full open-sourcing, makes consistent evaluation standards equally available to researchers and oversight bodies.
\paragraph{Negative impacts and mitigations} Agents that operate as intended but produce subtly incorrect outputs, for instance, hallucinated audit corrections or overconfident investment ratings, could mislead users into materially harmful decisions; we therefore release \FAB strictly as an evaluation harness, with no pretrained weights or deployable systems. Stronger agents identified through \FAB could also be misused to generate misleading market commentary or to amplify capital concentration among actors with privileged API access; open-sourcing the full benchmark and evaluation code is intended to make these capabilities and their failure modes auditable rather than proprietary. Finally, our coverage of only large-cap US equities and English-language disclosures may bias measured capability toward well-resourced market segments, and we encourage downstream extensions to non-US markets and underrepresented investor profiles.

\section*{Use of AI assistants}

AI coding assistants were used substantially to implement the benchmark infrastructure, the harness adapters, and the evaluation and analysis code, and AI tools were used to improve the clarity of the English prose and to support LaTeX formatting. LLMs are also part of the method itself, as the evaluated backbones, the rubric generators, and the report-quality verifiers (Appendix~\ref{sec:sec_evaluation_rubrics}). The authors designed the study, reviewed and tested all AI-assisted code and text, checked the reported results against the run records, and take full responsibility for the content of the paper.

% \section*{Acknowledgments}

\bibliographystyle{plainnat}
\bibliography{custom}
\appendix

\clearpage
\section{Extended Related Work}
\label{sec:sec_relatedwork}

\begin{table}[!b]
\centering
\footnotesize
\renewcommand{\arraystretch}{0.90}
\setlength{\tabcolsep}{6pt}

\begin{tabularx}{\linewidth}{l|*{4}{>{\centering\arraybackslash}X}|*{2}{>{\centering\arraybackslash}X}c}
\toprule
& \multicolumn{4}{c|}{\textbf{Financial Workflow Coverage}}
& \multicolumn{3}{c}{\textbf{Evaluation Design}} \\

\textbf{Benchmark}
& \textbf{Trade}
& \textbf{Hedge}
& \textbf{Insight}
& \textbf{Audit}
& \textbf{Intact}
& \textbf{E2E}
& \textbf{Cross-WF} \\
\midrule

\multicolumn{8}{@{}l}{
\footnotesize\textbf{\textit{General Agent Benchmarks}}} \\

AstaBench~\citep{bragg2025astabenchrigorousbenchmarkingai}
& \xmark & \xmark & \xmark & \xmark
& \cmark & \xmark & \xmark \\

General Agent Eval~\citep{bandel2026generalagentevaluation}
& \xmark & \xmark & \xmark & \xmark
& \cmark & \xmark & \xmark \\

\addlinespace[1pt]
\multicolumn{8}{@{}l}{
\footnotesize\textbf{\textit{Static Financial Benchmarks}}} \\

FinBen~\citep{NEURIPS2024_adb1d9fa}
& \cmark & \xmark & \xmark & \xmark
& \xmark & \xmark & \xmark \\

MultiFinBen~\citep{
peng2025multifinbenbenchmarkinglargelanguage}
& \cmark & \xmark & \xmark & \xmark
& \xmark & \xmark & \xmark \\

FinAuditing~\citep{
wang2026finauditingfinancialtaxonomystructuredmultidocument}
& \xmark & \xmark & \xmark & \cmark
& \xmark & \xmark & \xmark \\

\addlinespace[1pt]
\multicolumn{8}{@{}l}{
\footnotesize\textbf{\textit{Financial Agent Benchmarks}}} \\

FinRetrieval~\citep{kim2026finretrieval}
& \xmark & \xmark & \cmark & \xmark
& \cmark & \xmark & \xmark \\

FinMCP-Bench~\citep{qwen2025finmcp}
& \xmark & \xmark & \xmark & \xmark
& \xmark & \xmark & \xmark \\

Finance Agent Bench~\citep{bigeard2025fab}
& \xmark & \xmark & \xmark & \xmark
& \xmark & \xmark & \xmark \\

InvestorBench~\citep{li2025investorbench}
& \cmark & \xmark & \xmark & \xmark
& \xmark & \cmark & \xmark \\

StockBench~\citep{chen2025stockbench}
& \cmark & \xmark & \xmark & \xmark
& \xmark & \cmark & \xmark \\

Agent Market Arena~\citep{
qian2025agentstradelivemultimarket}
& \cmark & \xmark & \xmark & \xmark
& \cmark & \cmark & \xmark \\

Moira~\citep{
giannouris2026moiralanguagedrivenhierarchicalreinforcement}
& \xmark & \cmark & \xmark & \xmark
& \xmark & \xmark & \xmark \\

FinDeepResearch~\citep{
zhu2026findeepresearchevaluatingdeepresearch}
& \xmark & \xmark & \cmark & \xmark
& \cmark & \xmark & \xmark \\

FinDeepForecast~\citep{
li2026findeepforecastlivemultiagentbenchmarking}
& \xmark & \xmark & \xmark & \xmark
& \cmark & \xmark & \xmark \\

FINCH~\citep{dong2025finch}
& \xmark & \xmark & \cmark & \xmark
& \cmark & \xmark & \xmark \\

QFBench~\citep{quantitativefinancebench2026}
& \xmark & \xmark & \xmark & \xmark
& \cmark & \xmark & \xmark \\

\midrule
\rowcolor{lightblue}
\textbf{\FAB~(ours)}
& \cmark & \cmark & \cmark & \cmark
& \cmark & \cmark & \cmark \\

\bottomrule
\end{tabularx}

\caption{
Per-benchmark workflow coverage and evaluation design (compact version: Table~\ref{tab:benchmark_comparison}).
\textbf{Intact}: native control of independently designed agents is preserved;
\textbf{E2E}: from information acquisition to a committed outcome;
\textbf{Cross-WF}: matched architecture--backbone configurations across heterogeneous workflows.
}
\label{tab:benchmark_comparison_full}
\end{table}

\paragraph{Financial LLM benchmarks}
A large body of work evaluates the capability of large language models on financial tasks. 
These benchmarks typically focus on fixed-input settings, where models
answer questions over financial documents without controlling the
information-acquisition or execution process.
Early datasets focus on question answering and numerical reasoning. 
FinQA~\citep{DBLP:journals/corr/abs-2109-00122} evaluates numerical reasoning over financial reports, while DocFinQA~\citep{reddy2024docfinqa} studies long-context reasoning over document-level financial data. 
FinanceBench~\citep{islam2023financebenchnewbenchmarkfinancial} evaluates financial question answering grounded in corporate filings.
FinChain~\citep{xie2025finchain} further explores multi-step reasoning and verification.
More recent benchmarks broaden the evaluation scope. 
FinBen~\citep{NEURIPS2024_adb1d9fa}, FinReason~\cite{qian2025fino1}, OpenFinLLM~\citep{huang2025openfinllmsopenmultimodallarge}, and MultiFinBen~\citep{peng2025multifinbenbenchmarkinglargelanguage} introduce tasks covering reasoning, forecasting, and information extraction across financial documents, while XFinBench~\citep{zhang-etal-2025-xfinbench} focuses on complex financial problem solving with numerical reasoning and table understanding.
While useful for assessing LLM financial capabilities, these works mainly evaluate fixed-input tasks and do not test agentic abilities like executing multi-step financial workflows with tool use and sequential decision-making.

\paragraph{General agent benchmarks}
Alongside financial LLM benchmarks, a growing body of work studies the evaluation of general-purpose AI agents. 
However, these benchmarks do not reproduce the domain-specific states,
commitments, and outcome criteria that define financial workflows.
GTA~\citep{Wang2024GTA} evaluates tool-using agents through tasks involving human-written queries and visual inputs such as screenshots and tables. 
AgentBench~\citep{li2026benchmarktesttimescalinggeneral} provides a multi-domain benchmark with open-ended tasks and test-time scaling settings. 
Other benchmarks such as AstaBench~\citep{bragg2025astabenchrigorousbenchmarkingai} and GAIA~\citep{mialon2024gaia} introduce realistic environments and standardized evaluation procedures for assessing agent reasoning, tool use, and task execution, while General Agent Evaluation~\citep{bandel2026generalagentevaluation} proposes a unified protocol and agentic framework for benchmarking.

\paragraph{Financial agent benchmarks}
Recent work has begun to benchmark financial AI agents that reason, act, and interact with tools in finance-specific settings. 
However, these works typically either specialize in one financial
workflow or organize heterogeneous tasks as separate evaluation
settings, rather than examining the same agent systems across
end-to-end workflows with different execution requirements.
Early efforts target NLP-level capabilities such as retrieval, extraction, and tool use over financial documents~\citep{kim2026finretrieval, choi2025finagentbench, qwen2025finmcp}, establishing important baselines for information access in financial settings.
Subsequent work targets specialized financial tasks, particularly trading and market decision-making~\citep{li2025investorbench, chen2025stockbench, qian2025agentstradelivemultimarket}, introducing interactive evaluation and sequential decision-making in realistic market environments.
More recent benchmarks attempt broader financial analysis spanning multiple task types with tool access and multi-dimensional evaluation~\citep{bigeard2025fab, dong2025finch, quantitativefinancebench2026}.

\section{Multi-Run Evaluation}
\label{app:multirun}

Agent trajectories are stochastic: the same configuration can retrieve different evidence, take different actions, or fail to commit in one run but not another. We therefore run every agent--backbone configuration three times independently on each workflow, with identical environments, task instances, instructions, and interaction limits; only stochastic decoding and agent execution differ across runs. All metrics are computed within each run and then aggregated across runs, so that, for example, a Sharpe ratio is computed on each run's own return path rather than on an averaged path. Table~\ref{tab:results} reports the mean over the three runs, and Tables~\ref{tab:results_std}--\ref{tab:results_std_openclaw} report the mean together with the sample standard deviation ($n{=}3$), one table per agent. Decision steps without a committed submission are retained rather than excluded: they are replayed as \textsc{Hold} in Trading and Hedging, leave the position flat in Market Insights, and count as structural errors in Auditing. For Hedging, PAIR reports the most frequently selected pair and the number of runs that selected it.

% generated by analysis/build_results_table.py -- do not edit by hand
% needs table/tab_macros.tex (\bestup \secondup \bestdown \seconddown) and adjustbox
%
\begin{table*}[p]
\centering
\renewcommand{\arraystretch}{1.15}
\setlength{\tabcolsep}{3pt}
\begin{adjustbox}{max width=\textwidth, max totalheight=0.90\textheight}
\begin{tabular}{@{}lll|c|cccccc@{}}
\toprule
\multirow{2}{*}{\rotatebox{90}{\textbf{Task}}} & \multirow{2}{*}{\rotatebox{90}{\textbf{Stock}}} & \multirow{2}{*}{\textbf{Metric}}
& \textbf{BL} & \multicolumn{6}{c}{\textbf{ReAct Agent}} \\
\cmidrule(lr){4-4} \cmidrule(lr){5-10}
& & & B\&H & sonnet & gpt & ds32 & ds4p & qwen9 & qwen4 \\
\midrule
\multirow{12}{*}{\rotatebox{90}{Trading}}
& \multirow{3}{*}{\rotatebox{90}{MSFT}} & CR$\uparrow$ & -23.41 & -4.23\,$\pm$\,5.22 & 7.31\,$\pm$\,5.97 & -11.30\,$\pm$\,7.88 & 1.30\,$\pm$\,3.05 & -16.80\,$\pm$\,10.27 & -18.00\,$\pm$\,4.39 \\
& & SR$\uparrow$ & -3.50 & -0.51\,$\pm$\,0.79 & 1.19\,$\pm$\,0.71 & -1.65\,$\pm$\,1.28 & 0.31\,$\pm$\,0.45 & -3.08\,$\pm$\,2.06 & -3.36\,$\pm$\,0.76 \\
& & MDD$\downarrow$ & 25.12 & 10.70\,$\pm$\,0.27 & \seconddown{8.21}\,$\pm$\,0.89 & 18.24\,$\pm$\,2.87 & 13.35\,$\pm$\,3.47 & 18.38\,$\pm$\,9.56 & 18.82\,$\pm$\,3.90 \\
\cmidrule(lr){2-10}
& \multirow{3}{*}{\rotatebox{90}{TSLA}} & CR$\uparrow$ & -18.80 & -11.59\,$\pm$\,1.31 & -0.36\,$\pm$\,0.93 & 8.61\,$\pm$\,16.13 & -10.25\,$\pm$\,5.07 & -6.45\,$\pm$\,5.77 & -5.52\,$\pm$\,2.98 \\
& & SR$\uparrow$ & -2.60 & -1.58\,$\pm$\,0.22 & 0.10\,$\pm$\,0.13 & \secondup{1.33}\,$\pm$\,2.25 & -1.43\,$\pm$\,0.87 & -1.22\,$\pm$\,1.22 & -1.94\,$\pm$\,1.18 \\
& & MDD$\downarrow$ & 21.03 & 17.35\,$\pm$\,0.82 & 14.06\,$\pm$\,2.02 & 9.68\,$\pm$\,3.25 & 20.09\,$\pm$\,2.37 & 12.25\,$\pm$\,3.76 & 7.84\,$\pm$\,1.02 \\
\cmidrule(lr){2-10}
& \multirow{3}{*}{\rotatebox{90}{AAPL}} & CR$\uparrow$ & -6.70 & -0.92\,$\pm$\,2.77 & -3.76\,$\pm$\,4.89 & 0.90\,$\pm$\,0.58 & 4.35\,$\pm$\,1.99 & 0.84\,$\pm$\,11.53 & -5.21\,$\pm$\,7.07 \\
& & SR$\uparrow$ & -1.03 & -0.05\,$\pm$\,0.48 & -0.54\,$\pm$\,0.85 & 0.27\,$\pm$\,0.10 & 0.90\,$\pm$\,0.33 & 0.35\,$\pm$\,2.50 & -1.21\,$\pm$\,1.81 \\
& & MDD$\downarrow$ & 10.79 & 15.66\,$\pm$\,0.82 & 12.36\,$\pm$\,0.27 & 9.86\,$\pm$\,2.04 & 9.28\,$\pm$\,2.06 & 10.90\,$\pm$\,6.80 & 11.56\,$\pm$\,2.43 \\
\cmidrule(lr){2-10}
& \multirow{3}{*}{\rotatebox{90}{NVDA}} & CR$\uparrow$ & -8.13 & -21.45\,$\pm$\,5.02 & -10.68\,$\pm$\,2.65 & -5.95\,$\pm$\,4.22 & -19.16\,$\pm$\,2.90 & -6.64\,$\pm$\,11.43 & -4.34\,$\pm$\,7.62 \\
& & SR$\uparrow$ & -0.86 & -3.19\,$\pm$\,1.10 & -1.31\,$\pm$\,0.35 & -0.67\,$\pm$\,0.57 & -2.55\,$\pm$\,0.37 & -0.91\,$\pm$\,1.70 & -0.95\,$\pm$\,1.87 \\
& & MDD$\downarrow$ & 14.13 & 22.23\,$\pm$\,4.97 & 14.39\,$\pm$\,0.90 & 15.64\,$\pm$\,3.68 & 21.48\,$\pm$\,4.80 & 13.68\,$\pm$\,7.12 & 11.70\,$\pm$\,4.08 \\
\midrule
\multirow{4}{*}{\rotatebox{90}{Hedging}}
& & PAIR & --- & \shortstack[c]{META\\MSFT\\(1/3)} & \shortstack[c]{MSFT\\NVDA\\(2/3)} & \shortstack[c]{NVDA\\MSFT\\(3/3)} & \shortstack[c]{NVDA\\MSFT\\(3/3)} & \shortstack[c]{MSFT\\NVDA\\(2/3)} & \shortstack[c]{MSFT\\NVDA\\(2/3)} \\
& & CR$\uparrow$ & --- & -4.54\,$\pm$\,7.75 & 2.06\,$\pm$\,4.34 & -8.67\,$\pm$\,1.60 & -8.65\,$\pm$\,0.95 & -8.79\,$\pm$\,7.22 & -8.31\,$\pm$\,6.85 \\
& & SR$\uparrow$ & --- & -1.29\,$\pm$\,1.90 & 0.89\,$\pm$\,1.81 & -2.10\,$\pm$\,0.42 & -1.95\,$\pm$\,0.24 & -1.77\,$\pm$\,1.38 & -1.88\,$\pm$\,1.62 \\
& & MDD$\downarrow$ & --- & 9.19\,$\pm$\,1.67 & \bestdown{2.90}\,$\pm$\,1.03 & 11.49\,$\pm$\,0.66 & 12.23\,$\pm$\,0.06 & 13.48\,$\pm$\,4.18 & 12.00\,$\pm$\,5.21 \\
\midrule
\multirow{16}{*}{\rotatebox{90}{Market Insights}}
& \multirow{4}{*}{\rotatebox{90}{MSFT}} & Score$\uparrow$ & --- & 8.16\,$\pm$\,0.03 & 8.36\,$\pm$\,0.04 & 6.67\,$\pm$\,0.23 & 8.51\,$\pm$\,0.19 & 4.89\,$\pm$\,2.64 & 6.41\,$\pm$\,0.74 \\
& & CR$\uparrow$ & -21.74 & -8.00\,$\pm$\,1.51 & \bestup{-2.74}\,$\pm$\,0.46 & -7.11\,$\pm$\,1.88 & -8.17\,$\pm$\,5.20 & -5.31\,$\pm$\,5.53 & -6.91\,$\pm$\,4.12 \\
& & SR$\uparrow$ & -3.25 & -2.70\,$\pm$\,0.41 & -3.21\,$\pm$\,0.52 & -2.41\,$\pm$\,0.41 & -3.35\,$\pm$\,1.08 & -1.94\,$\pm$\,2.24 & -2.19\,$\pm$\,0.78 \\
& & MDD$\downarrow$ & 24.93 & 9.04\,$\pm$\,1.29 & \bestdown{3.48}\,$\pm$\,0.43 & 9.34\,$\pm$\,1.48 & 9.17\,$\pm$\,5.58 & 7.56\,$\pm$\,4.82 & 8.61\,$\pm$\,3.27 \\
\cmidrule(lr){2-10}
& \multirow{4}{*}{\rotatebox{90}{TSLA}} & Score$\uparrow$ & --- & 7.98\,$\pm$\,0.08 & 8.23\,$\pm$\,0.04 & 7.27\,$\pm$\,0.33 & 8.04\,$\pm$\,0.42 & 4.57\,$\pm$\,3.68 & 6.44\,$\pm$\,1.46 \\
& & CR$\uparrow$ & -17.02 & 0.00\,$\pm$\,0.00 & 1.55\,$\pm$\,1.92 & 0.24\,$\pm$\,0.41 & 0.41\,$\pm$\,0.54 & -1.39\,$\pm$\,1.93 & -0.20\,$\pm$\,1.73 \\
& & SR$\uparrow$ & -2.35 & 0.00\,$\pm$\,0.00 & \bestup{0.77}\,$\pm$\,0.63 & 0.26\,$\pm$\,0.45 & 0.19\,$\pm$\,0.18 & -0.85\,$\pm$\,1.20 & -0.12\,$\pm$\,1.30 \\
& & MDD$\downarrow$ & 19.70 & 0.00\,$\pm$\,0.00 & 3.48\,$\pm$\,1.74 & \bestdown{0.51}\,$\pm$\,0.89 & 3.14\,$\pm$\,0.98 & 2.15\,$\pm$\,1.90 & 1.81\,$\pm$\,1.58 \\
\cmidrule(lr){2-10}
& \multirow{4}{*}{\rotatebox{90}{AAPL}} & Score$\uparrow$ & --- & 8.43\,$\pm$\,0.18 & 8.79\,$\pm$\,0.07 & 7.08\,$\pm$\,0.77 & 8.94\,$\pm$\,0.13 & 5.56\,$\pm$\,3.84 & 6.86\,$\pm$\,0.64 \\
& & CR$\uparrow$ & -6.14 & -1.15\,$\pm$\,1.35 & -0.28\,$\pm$\,0.00 & -1.14\,$\pm$\,2.68 & -0.68\,$\pm$\,1.38 & -1.44\,$\pm$\,6.35 & -4.63\,$\pm$\,4.66 \\
& & SR$\uparrow$ & -0.94 & -0.43\,$\pm$\,0.56 & -0.13\,$\pm$\,0.00 & -0.40\,$\pm$\,1.03 & -0.24\,$\pm$\,0.55 & 0.02\,$\pm$\,2.48 & -1.53\,$\pm$\,1.65 \\
& & MDD$\downarrow$ & 10.79 & 6.03\,$\pm$\,0.83 & \bestdown{3.60}\,$\pm$\,0.00 & 5.02\,$\pm$\,0.69 & 4.90\,$\pm$\,1.13 & 6.82\,$\pm$\,4.43 & 7.83\,$\pm$\,2.23 \\
\cmidrule(lr){2-10}
& \multirow{4}{*}{\rotatebox{90}{NVDA}} & Score$\uparrow$ & --- & 8.34\,$\pm$\,0.14 & 8.62\,$\pm$\,0.03 & 7.29\,$\pm$\,0.50 & 8.96\,$\pm$\,0.04 & 6.55\,$\pm$\,1.69 & 7.58\,$\pm$\,0.41 \\
& & CR$\uparrow$ & -9.05 & -3.52\,$\pm$\,1.36 & -3.65\,$\pm$\,1.00 & -2.78\,$\pm$\,4.06 & -3.71\,$\pm$\,0.95 & -5.28\,$\pm$\,5.12 & -5.42\,$\pm$\,2.86 \\
& & SR$\uparrow$ & -1.00 & -0.80\,$\pm$\,0.33 & -1.06\,$\pm$\,0.16 & -0.66\,$\pm$\,1.18 & -0.93\,$\pm$\,0.26 & -1.02\,$\pm$\,0.69 & -1.10\,$\pm$\,0.51 \\
& & MDD$\downarrow$ & 14.13 & 6.49\,$\pm$\,1.24 & \bestdown{4.84}\,$\pm$\,0.38 & 6.71\,$\pm$\,2.15 & 5.94\,$\pm$\,1.12 & 8.98\,$\pm$\,5.97 & 8.89\,$\pm$\,3.42 \\
\midrule
\multirow{4}{*}{\rotatebox{90}{Auditing}}
& & ACC$\uparrow$ & --- & 49.75\,$\pm$\,2.28 & 28.86\,$\pm$\,3.11 & 15.92\,$\pm$\,2.28 & 49.25\,$\pm$\,2.99 & 28.36\,$\pm$\,12.93 & 28.86\,$\pm$\,10.80 \\
& & SER$\downarrow$ & --- & 0.00\,$\pm$\,0.00 & 7.96\,$\pm$\,3.11 & 59.20\,$\pm$\,1.72 & 12.94\,$\pm$\,1.72 & 28.86\,$\pm$\,16.44 & 21.39\,$\pm$\,12.52 \\
& & EER$\downarrow$ & --- & 2.49\,$\pm$\,0.86 & 0.00\,$\pm$\,0.00 & 0.00\,$\pm$\,0.00 & 0.00\,$\pm$\,0.00 & 0.00\,$\pm$\,0.00 & 0.00\,$\pm$\,0.00 \\
& & CER$\downarrow$ & --- & 47.76\,$\pm$\,2.59 & 63.18\,$\pm$\,3.11 & 24.88\,$\pm$\,0.86 & 37.81\,$\pm$\,3.45 & 42.79\,$\pm$\,3.76 & 49.75\,$\pm$\,2.28 \\
\bottomrule
\end{tabular}
\end{adjustbox}
\caption{Performance of ReAct Agent on the \FAB benchmark: mean\,$\pm$\,std over three independent runs; metrics are computed per run, then aggregated. Metrics, backbones, and notation as in Table~\ref{tab:results}; PAIR is the most frequently selected Hedging pair, with the number of runs selecting it; best and second-best means are ranked among all 30 configurations (Tables~\ref{tab:results_std}--\ref{tab:results_std_openclaw}).}
\label{tab:results_std}
\end{table*}

\begin{table*}[p]
\centering
\renewcommand{\arraystretch}{1.15}
\setlength{\tabcolsep}{3pt}
\begin{adjustbox}{max width=\textwidth, max totalheight=0.90\textheight}
\begin{tabular}{@{}lll|c|cccccc@{}}
\toprule
\multirow{2}{*}{\rotatebox{90}{\textbf{Task}}} & \multirow{2}{*}{\rotatebox{90}{\textbf{Stock}}} & \multirow{2}{*}{\textbf{Metric}}
& \textbf{BL} & \multicolumn{6}{c}{\textbf{Claude Code}} \\
\cmidrule(lr){4-4} \cmidrule(lr){5-10}
& & & B\&H & sonnet & gpt & ds32 & ds4p & qwen9 & qwen4 \\
\midrule
\multirow{12}{*}{\rotatebox{90}{Trading}}
& \multirow{3}{*}{\rotatebox{90}{MSFT}} & CR$\uparrow$ & -23.41 & -1.15\,$\pm$\,8.17 & \secondup{17.41}\,$\pm$\,27.82 & -7.15\,$\pm$\,12.31 & -5.85\,$\pm$\,4.84 & -12.87\,$\pm$\,6.12 & -10.43\,$\pm$\,7.35 \\
& & SR$\uparrow$ & -3.50 & 0.02\,$\pm$\,1.27 & \secondup{2.20}\,$\pm$\,3.23 & -1.08\,$\pm$\,2.27 & -0.74\,$\pm$\,0.76 & -2.11\,$\pm$\,1.12 & -1.89\,$\pm$\,1.48 \\
& & MDD$\downarrow$ & 25.12 & 9.63\,$\pm$\,1.12 & 8.42\,$\pm$\,4.28 & 13.77\,$\pm$\,6.07 & 15.42\,$\pm$\,1.18 & 17.49\,$\pm$\,5.17 & 14.38\,$\pm$\,4.01 \\
\cmidrule(lr){2-10}
& \multirow{3}{*}{\rotatebox{90}{TSLA}} & CR$\uparrow$ & -18.80 & -14.08\,$\pm$\,4.28 & \secondup{9.63}\,$\pm$\,12.75 & 1.43\,$\pm$\,6.63 & -5.19\,$\pm$\,6.28 & -6.02\,$\pm$\,8.19 & -5.30\,$\pm$\,5.10 \\
& & SR$\uparrow$ & -2.60 & -1.95\,$\pm$\,0.73 & \secondup{1.33}\,$\pm$\,1.59 & 0.25\,$\pm$\,1.09 & -0.63\,$\pm$\,0.92 & -1.78\,$\pm$\,2.51 & -1.91\,$\pm$\,2.00 \\
& & MDD$\downarrow$ & 21.03 & 20.10\,$\pm$\,4.33 & 10.55\,$\pm$\,4.85 & 10.64\,$\pm$\,4.87 & 17.06\,$\pm$\,1.25 & 10.81\,$\pm$\,5.09 & \seconddown{7.53}\,$\pm$\,3.14 \\
\cmidrule(lr){2-10}
& \multirow{3}{*}{\rotatebox{90}{AAPL}} & CR$\uparrow$ & -6.70 & 1.08\,$\pm$\,2.29 & -6.93\,$\pm$\,2.34 & 6.47\,$\pm$\,3.95 & \secondup{8.75}\,$\pm$\,8.87 & 2.26\,$\pm$\,2.79 & 1.05\,$\pm$\,2.99 \\
& & SR$\uparrow$ & -1.03 & 0.30\,$\pm$\,0.38 & -1.12\,$\pm$\,0.44 & 1.43\,$\pm$\,0.90 & \bestup{1.76}\,$\pm$\,1.71 & 0.66\,$\pm$\,0.68 & 0.65\,$\pm$\,1.04 \\
& & MDD$\downarrow$ & 10.79 & 11.89\,$\pm$\,2.42 & 13.90\,$\pm$\,3.62 & 7.02\,$\pm$\,0.43 & 6.82\,$\pm$\,1.72 & 7.23\,$\pm$\,2.70 & \bestdown{4.85}\,$\pm$\,3.87 \\
\cmidrule(lr){2-10}
& \multirow{3}{*}{\rotatebox{90}{NVDA}} & CR$\uparrow$ & -8.13 & -24.89\,$\pm$\,4.46 & -6.35\,$\pm$\,10.97 & -11.82\,$\pm$\,12.14 & -15.65\,$\pm$\,11.09 & -8.32\,$\pm$\,6.38 & -6.08\,$\pm$\,6.39 \\
& & SR$\uparrow$ & -0.86 & -3.98\,$\pm$\,0.69 & -0.69\,$\pm$\,1.44 & -1.79\,$\pm$\,2.14 & -2.11\,$\pm$\,1.65 & -1.40\,$\pm$\,1.25 & -1.06\,$\pm$\,1.13 \\
& & MDD$\downarrow$ & 14.13 & 25.64\,$\pm$\,4.42 & 14.02\,$\pm$\,2.96 & 17.43\,$\pm$\,8.49 & 20.08\,$\pm$\,6.38 & 13.76\,$\pm$\,4.17 & 12.66\,$\pm$\,6.06 \\
\midrule
\multirow{4}{*}{\rotatebox{90}{Hedging}}
& & PAIR & --- & \shortstack[c]{NVDA\\TSLA\\(1/3)} & \shortstack[c]{META\\MSFT\\(3/3)} & \shortstack[c]{NVDA\\MSFT\\(3/3)} & \shortstack[c]{NVDA\\MSFT\\(2/3)} & \shortstack[c]{MSFT\\AAPL\\(1/3)} & \shortstack[c]{GOOGL\\TSLA\\(1/3)} \\
& & CR$\uparrow$ & --- & -0.89\,$\pm$\,11.98 & 1.82\,$\pm$\,4.69 & -7.51\,$\pm$\,1.52 & -9.54\,$\pm$\,1.44 & \bestup{7.01}\,$\pm$\,1.08 & 3.94\,$\pm$\,2.31 \\
& & SR$\uparrow$ & --- & -0.22\,$\pm$\,2.93 & 1.00\,$\pm$\,1.84 & -1.74\,$\pm$\,0.36 & -2.05\,$\pm$\,0.57 & \bestup{1.71}\,$\pm$\,0.37 & 1.09\,$\pm$\,0.65 \\
& & MDD$\downarrow$ & --- & 8.47\,$\pm$\,6.45 & 7.88\,$\pm$\,5.15 & 11.88\,$\pm$\,0.77 & 12.39\,$\pm$\,0.20 & 4.74\,$\pm$\,1.11 & 6.32\,$\pm$\,2.42 \\
\midrule
\multirow{16}{*}{\rotatebox{90}{Market Insights}}
& \multirow{4}{*}{\rotatebox{90}{MSFT}} & Score$\uparrow$ & --- & 8.14\,$\pm$\,0.04 & 8.28\,$\pm$\,0.19 & 3.72\,$\pm$\,1.32 & \bestup{8.62}\,$\pm$\,0.09 & 6.58\,$\pm$\,1.18 & 5.50\,$\pm$\,0.86 \\
& & CR$\uparrow$ & -21.74 & -7.71\,$\pm$\,3.75 & -3.91\,$\pm$\,1.90 & -3.97\,$\pm$\,5.22 & -7.71\,$\pm$\,3.54 & -5.68\,$\pm$\,2.00 & \secondup{-2.83}\,$\pm$\,1.42 \\
& & SR$\uparrow$ & -3.25 & -2.64\,$\pm$\,1.68 & -2.96\,$\pm$\,0.48 & \bestup{-0.83}\,$\pm$\,3.18 & -2.67\,$\pm$\,0.20 & -3.58\,$\pm$\,1.11 & -1.21\,$\pm$\,0.59 \\
& & MDD$\downarrow$ & 24.93 & 10.03\,$\pm$\,2.22 & \seconddown{4.63}\,$\pm$\,1.89 & 5.40\,$\pm$\,5.12 & 9.49\,$\pm$\,4.63 & 7.29\,$\pm$\,3.56 & 5.83\,$\pm$\,0.32 \\
\cmidrule(lr){2-10}
& \multirow{4}{*}{\rotatebox{90}{TSLA}} & Score$\uparrow$ & --- & 8.08\,$\pm$\,0.27 & \secondup{8.54}\,$\pm$\,0.07 & 4.71\,$\pm$\,0.93 & \bestup{8.56}\,$\pm$\,0.02 & 6.47\,$\pm$\,1.21 & 5.47\,$\pm$\,1.49 \\
& & CR$\uparrow$ & -17.02 & 0.24\,$\pm$\,0.41 & -0.10\,$\pm$\,0.50 & -1.40\,$\pm$\,2.45 & 0.07\,$\pm$\,0.77 & -0.73\,$\pm$\,2.06 & -0.17\,$\pm$\,2.27 \\
& & SR$\uparrow$ & -2.35 & 0.26\,$\pm$\,0.45 & -0.02\,$\pm$\,0.26 & -1.59\,$\pm$\,2.35 & 0.09\,$\pm$\,0.39 & -0.44\,$\pm$\,1.39 & 0.15\,$\pm$\,1.39 \\
& & MDD$\downarrow$ & 19.70 & \bestdown{0.51}\,$\pm$\,0.89 & 4.45\,$\pm$\,0.79 & 2.84\,$\pm$\,0.39 & 3.81\,$\pm$\,1.02 & 2.43\,$\pm$\,2.53 & 2.61\,$\pm$\,1.21 \\
\cmidrule(lr){2-10}
& \multirow{4}{*}{\rotatebox{90}{AAPL}} & Score$\uparrow$ & --- & 8.82\,$\pm$\,0.14 & 8.80\,$\pm$\,0.16 & 5.24\,$\pm$\,1.24 & 8.92\,$\pm$\,0.23 & 6.61\,$\pm$\,0.82 & 6.61\,$\pm$\,0.96 \\
& & CR$\uparrow$ & -6.14 & -0.94\,$\pm$\,3.57 & -0.07\,$\pm$\,0.20 & 1.19\,$\pm$\,6.10 & \secondup{1.36}\,$\pm$\,1.60 & -2.01\,$\pm$\,6.80 & 0.10\,$\pm$\,2.44 \\
& & SR$\uparrow$ & -0.94 & -0.36\,$\pm$\,1.47 & 0.00\,$\pm$\,0.12 & \bestup{1.03}\,$\pm$\,2.42 & 0.45\,$\pm$\,0.51 & -0.45\,$\pm$\,3.04 & 0.12\,$\pm$\,1.16 \\
& & MDD$\downarrow$ & 10.79 & 5.33\,$\pm$\,0.87 & 4.49\,$\pm$\,0.77 & 4.43\,$\pm$\,3.12 & 6.81\,$\pm$\,1.74 & 6.11\,$\pm$\,3.80 & 4.08\,$\pm$\,0.34 \\
\cmidrule(lr){2-10}
& \multirow{4}{*}{\rotatebox{90}{NVDA}} & Score$\uparrow$ & --- & 8.63\,$\pm$\,0.04 & 8.77\,$\pm$\,0.17 & 4.64\,$\pm$\,0.81 & 8.75\,$\pm$\,0.38 & 7.21\,$\pm$\,0.68 & 7.15\,$\pm$\,0.36 \\
& & CR$\uparrow$ & -9.05 & \secondup{-1.95}\,$\pm$\,0.00 & -5.18\,$\pm$\,1.24 & \bestup{0.28}\,$\pm$\,1.70 & -5.01\,$\pm$\,0.83 & -3.68\,$\pm$\,2.15 & -5.51\,$\pm$\,1.94 \\
& & SR$\uparrow$ & -1.00 & \secondup{-0.41}\,$\pm$\,0.00 & -1.60\,$\pm$\,0.47 & \bestup{0.17}\,$\pm$\,0.58 & -1.31\,$\pm$\,0.16 & -0.94\,$\pm$\,0.51 & -1.37\,$\pm$\,0.80 \\
& & MDD$\downarrow$ & 14.13 & \seconddown{5.06}\,$\pm$\,0.00 & 5.88\,$\pm$\,1.09 & 5.34\,$\pm$\,1.64 & 5.85\,$\pm$\,0.82 & 5.78\,$\pm$\,0.84 & 8.35\,$\pm$\,1.43 \\
\midrule
\multirow{4}{*}{\rotatebox{90}{Auditing}}
& & ACC$\uparrow$ & --- & 44.28\,$\pm$\,1.72 & 26.37\,$\pm$\,4.56 & 17.41\,$\pm$\,1.72 & \bestup{54.73}\,$\pm$\,3.76 & 13.93\,$\pm$\,0.86 & 25.37\,$\pm$\,0.00 \\
& & SER$\downarrow$ & --- & 0.00\,$\pm$\,0.00 & 23.38\,$\pm$\,1.72 & 40.80\,$\pm$\,3.76 & 5.47\,$\pm$\,1.72 & 1.49\,$\pm$\,1.49 & 4.48\,$\pm$\,3.95 \\
& & EER$\downarrow$ & --- & 3.48\,$\pm$\,0.86 & 1.99\,$\pm$\,0.86 & 0.00\,$\pm$\,0.00 & 0.00\,$\pm$\,0.00 & 0.50\,$\pm$\,0.86 & 1.99\,$\pm$\,0.86 \\
& & CER$\downarrow$ & --- & 52.24\,$\pm$\,1.49 & 48.26\,$\pm$\,3.76 & 41.79\,$\pm$\,3.95 & 39.80\,$\pm$\,2.28 & 84.08\,$\pm$\,1.72 & 68.16\,$\pm$\,4.56 \\
\bottomrule
\end{tabular}
\end{adjustbox}
\caption{Performance of Claude Code on the \FAB benchmark: mean\,$\pm$\,std over three runs; metrics and notation as in Table~\ref{tab:results_std}.}
\label{tab:results_std_claude_code}
\end{table*}

\begin{table*}[p]
\centering
\renewcommand{\arraystretch}{1.15}
\setlength{\tabcolsep}{3pt}
\begin{adjustbox}{max width=\textwidth, max totalheight=0.90\textheight}
\begin{tabular}{@{}lll|c|cccccc@{}}
\toprule
\multirow{2}{*}{\rotatebox{90}{\textbf{Task}}} & \multirow{2}{*}{\rotatebox{90}{\textbf{Stock}}} & \multirow{2}{*}{\textbf{Metric}}
& \textbf{BL} & \multicolumn{6}{c}{\textbf{Codex}} \\
\cmidrule(lr){4-4} \cmidrule(lr){5-10}
& & & B\&H & sonnet & gpt & ds32 & ds4p & qwen9 & qwen4 \\
\midrule
\multirow{12}{*}{\rotatebox{90}{Trading}}
& \multirow{3}{*}{\rotatebox{90}{MSFT}} & CR$\uparrow$ & -23.41 & 3.29\,$\pm$\,3.56 & -7.15\,$\pm$\,13.21 & -9.68\,$\pm$\,3.45 & -6.56\,$\pm$\,7.27 & -12.64\,$\pm$\,14.61 & -7.39\,$\pm$\,0.59 \\
& & SR$\uparrow$ & -3.50 & 0.59\,$\pm$\,0.50 & -1.00\,$\pm$\,2.11 & -1.31\,$\pm$\,0.56 & -0.82\,$\pm$\,1.09 & -2.08\,$\pm$\,2.66 & -1.50\,$\pm$\,0.31 \\
& & MDD$\downarrow$ & 25.12 & 9.11\,$\pm$\,0.00 & 14.99\,$\pm$\,6.49 & 17.37\,$\pm$\,1.63 & 18.07\,$\pm$\,1.30 & 16.38\,$\pm$\,10.21 & 10.34\,$\pm$\,2.65 \\
\cmidrule(lr){2-10}
& \multirow{3}{*}{\rotatebox{90}{TSLA}} & CR$\uparrow$ & -18.80 & -8.14\,$\pm$\,2.99 & -9.48\,$\pm$\,5.92 & 8.59\,$\pm$\,22.10 & -19.34\,$\pm$\,3.79 & -9.61\,$\pm$\,2.58 & -12.35\,$\pm$\,2.56 \\
& & SR$\uparrow$ & -2.60 & -0.99\,$\pm$\,0.44 & -1.23\,$\pm$\,0.89 & 1.09\,$\pm$\,3.01 & -2.95\,$\pm$\,0.67 & -1.68\,$\pm$\,0.56 & -2.95\,$\pm$\,0.81 \\
& & MDD$\downarrow$ & 21.03 & 16.18\,$\pm$\,2.34 & 18.83\,$\pm$\,1.11 & 15.51\,$\pm$\,7.20 & 24.79\,$\pm$\,0.87 & 15.70\,$\pm$\,1.52 & 14.59\,$\pm$\,0.44 \\
\cmidrule(lr){2-10}
& \multirow{3}{*}{\rotatebox{90}{AAPL}} & CR$\uparrow$ & -6.70 & 6.13\,$\pm$\,2.41 & 1.14\,$\pm$\,7.08 & 8.26\,$\pm$\,2.37 & 1.33\,$\pm$\,3.27 & -2.78\,$\pm$\,7.08 & -0.12\,$\pm$\,6.43 \\
& & SR$\uparrow$ & -1.03 & 1.12\,$\pm$\,0.38 & 0.27\,$\pm$\,1.19 & 1.54\,$\pm$\,0.39 & 0.37\,$\pm$\,0.58 & -0.52\,$\pm$\,1.55 & 0.23\,$\pm$\,1.70 \\
& & MDD$\downarrow$ & 10.79 & 12.55\,$\pm$\,1.52 & 12.93\,$\pm$\,1.84 & \seconddown{6.78}\,$\pm$\,1.47 & 10.76\,$\pm$\,3.37 & 8.87\,$\pm$\,3.92 & 6.91\,$\pm$\,4.26 \\
\cmidrule(lr){2-10}
& \multirow{3}{*}{\rotatebox{90}{NVDA}} & CR$\uparrow$ & -8.13 & -24.25\,$\pm$\,3.21 & -20.38\,$\pm$\,6.78 & -12.96\,$\pm$\,15.68 & -18.25\,$\pm$\,7.67 & -14.64\,$\pm$\,5.59 & \secondup{-1.81}\,$\pm$\,8.51 \\
& & SR$\uparrow$ & -0.86 & -3.57\,$\pm$\,0.58 & -2.68\,$\pm$\,1.09 & -1.83\,$\pm$\,2.42 & -2.40\,$\pm$\,1.15 & -2.20\,$\pm$\,1.09 & \secondup{-0.60}\,$\pm$\,2.02 \\
& & MDD$\downarrow$ & 14.13 & 25.01\,$\pm$\,3.18 & 21.30\,$\pm$\,6.69 & 18.92\,$\pm$\,9.59 & 21.27\,$\pm$\,4.15 & 20.01\,$\pm$\,4.65 & \seconddown{8.98}\,$\pm$\,2.82 \\
\midrule
\multirow{4}{*}{\rotatebox{90}{Hedging}}
& & PAIR & --- & \shortstack[c]{NVDA\\MSFT\\(3/3)} & \shortstack[c]{MSFT\\NVDA\\(2/3)} & \shortstack[c]{MSFT\\NVDA\\(2/3)} & \shortstack[c]{META\\MSFT\\(1/3)} & \shortstack[c]{AAPL\\MSFT\\(1/3)} & \shortstack[c]{GOOGL\\META\\(1/3)} \\
& & CR$\uparrow$ & --- & -9.88\,$\pm$\,2.99 & -0.12\,$\pm$\,5.07 & -7.50\,$\pm$\,5.55 & -4.64\,$\pm$\,4.71 & -5.96\,$\pm$\,3.84 & -2.51\,$\pm$\,4.69 \\
& & SR$\uparrow$ & --- & -2.56\,$\pm$\,0.84 & -0.10\,$\pm$\,1.89 & -1.71\,$\pm$\,1.48 & -0.88\,$\pm$\,1.05 & -1.20\,$\pm$\,0.87 & -0.49\,$\pm$\,1.07 \\
& & MDD$\downarrow$ & --- & 12.99\,$\pm$\,0.92 & \seconddown{3.88}\,$\pm$\,0.95 & 12.25\,$\pm$\,2.29 & 12.03\,$\pm$\,0.64 & 10.17\,$\pm$\,3.62 & 7.23\,$\pm$\,2.34 \\
\midrule
\multirow{16}{*}{\rotatebox{90}{Market Insights}}
& \multirow{4}{*}{\rotatebox{90}{MSFT}} & Score$\uparrow$ & --- & 8.21\,$\pm$\,0.09 & 8.32\,$\pm$\,0.22 & 3.66\,$\pm$\,0.51 & 8.43\,$\pm$\,0.17 & 7.33\,$\pm$\,0.12 & 7.00\,$\pm$\,0.19 \\
& & CR$\uparrow$ & -21.74 & -4.16\,$\pm$\,0.68 & -4.39\,$\pm$\,1.64 & -7.78\,$\pm$\,2.19 & -4.03\,$\pm$\,1.19 & -7.83\,$\pm$\,2.45 & -6.88\,$\pm$\,3.27 \\
& & SR$\uparrow$ & -3.25 & -2.31\,$\pm$\,1.22 & -3.12\,$\pm$\,0.74 & -2.81\,$\pm$\,0.70 & -1.45\,$\pm$\,0.55 & -2.87\,$\pm$\,0.94 & -2.34\,$\pm$\,0.95 \\
& & MDD$\downarrow$ & 24.93 & 5.88\,$\pm$\,1.31 & 5.60\,$\pm$\,2.48 & 9.39\,$\pm$\,3.14 & 6.86\,$\pm$\,1.41 & 9.04\,$\pm$\,2.21 & 9.20\,$\pm$\,2.46 \\
\cmidrule(lr){2-10}
& \multirow{4}{*}{\rotatebox{90}{TSLA}} & Score$\uparrow$ & --- & 8.12\,$\pm$\,0.14 & 8.08\,$\pm$\,0.10 & 4.88\,$\pm$\,1.29 & 8.10\,$\pm$\,0.14 & 7.27\,$\pm$\,0.06 & 6.81\,$\pm$\,0.20 \\
& & CR$\uparrow$ & -17.02 & 0.50\,$\pm$\,0.52 & \bestup{2.13}\,$\pm$\,3.39 & 0.57\,$\pm$\,0.51 & -0.74\,$\pm$\,1.12 & -1.58\,$\pm$\,2.29 & -1.53\,$\pm$\,3.66 \\
& & SR$\uparrow$ & -2.35 & 0.24\,$\pm$\,0.21 & \secondup{0.67}\,$\pm$\,0.94 & 0.46\,$\pm$\,0.41 & -0.50\,$\pm$\,0.87 & -0.71\,$\pm$\,0.94 & -0.43\,$\pm$\,2.15 \\
& & MDD$\downarrow$ & 19.70 & 2.27\,$\pm$\,2.11 & 3.96\,$\pm$\,0.96 & 1.62\,$\pm$\,1.66 & 4.86\,$\pm$\,0.65 & 3.60\,$\pm$\,1.22 & 5.37\,$\pm$\,4.67 \\
\cmidrule(lr){2-10}
& \multirow{4}{*}{\rotatebox{90}{AAPL}} & Score$\uparrow$ & --- & 8.56\,$\pm$\,0.11 & 8.59\,$\pm$\,0.10 & 4.75\,$\pm$\,1.06 & 8.60\,$\pm$\,0.11 & 7.55\,$\pm$\,0.06 & 7.27\,$\pm$\,0.20 \\
& & CR$\uparrow$ & -6.14 & -1.18\,$\pm$\,2.53 & -0.84\,$\pm$\,0.78 & -2.05\,$\pm$\,2.40 & 0.91\,$\pm$\,2.16 & -0.13\,$\pm$\,1.74 & 0.18\,$\pm$\,2.34 \\
& & SR$\uparrow$ & -0.94 & -0.48\,$\pm$\,1.12 & -0.39\,$\pm$\,0.35 & -0.69\,$\pm$\,0.69 & 0.53\,$\pm$\,1.09 & 0.08\,$\pm$\,0.56 & 0.32\,$\pm$\,1.30 \\
& & MDD$\downarrow$ & 10.79 & 4.95\,$\pm$\,0.50 & 4.07\,$\pm$\,0.81 & 6.02\,$\pm$\,2.67 & \seconddown{4.04}\,$\pm$\,1.32 & 6.19\,$\pm$\,1.64 & 4.42\,$\pm$\,1.39 \\
\cmidrule(lr){2-10}
& \multirow{4}{*}{\rotatebox{90}{NVDA}} & Score$\uparrow$ & --- & 8.71\,$\pm$\,0.23 & 8.65\,$\pm$\,0.14 & 2.85\,$\pm$\,0.46 & 8.66\,$\pm$\,0.20 & 7.55\,$\pm$\,0.25 & 7.34\,$\pm$\,0.11 \\
& & CR$\uparrow$ & -9.05 & -3.44\,$\pm$\,1.43 & -3.94\,$\pm$\,0.26 & -4.24\,$\pm$\,2.10 & -3.97\,$\pm$\,0.30 & -5.17\,$\pm$\,3.20 & -6.25\,$\pm$\,1.84 \\
& & SR$\uparrow$ & -1.00 & -0.81\,$\pm$\,0.37 & -1.09\,$\pm$\,0.09 & -2.02\,$\pm$\,0.64 & -1.03\,$\pm$\,0.08 & -1.49\,$\pm$\,1.08 & -1.70\,$\pm$\,0.63 \\
& & MDD$\downarrow$ & 14.13 & 5.94\,$\pm$\,1.52 & \seconddown{5.06}\,$\pm$\,0.00 & 5.09\,$\pm$\,1.52 & 5.77\,$\pm$\,1.24 & 6.94\,$\pm$\,1.77 & 8.16\,$\pm$\,1.52 \\
\midrule
\multirow{4}{*}{\rotatebox{90}{Auditing}}
& & ACC$\uparrow$ & --- & 45.27\,$\pm$\,0.86 & 33.33\,$\pm$\,3.11 & 42.29\,$\pm$\,4.56 & 50.75\,$\pm$\,0.00 & 28.36\,$\pm$\,2.59 & 29.35\,$\pm$\,2.28 \\
& & SER$\downarrow$ & --- & 0.50\,$\pm$\,0.86 & 1.99\,$\pm$\,2.28 & 0.50\,$\pm$\,0.86 & 1.49\,$\pm$\,1.49 & 0.00\,$\pm$\,0.00 & 1.49\,$\pm$\,1.49 \\
& & EER$\downarrow$ & --- & 1.99\,$\pm$\,1.72 & 1.00\,$\pm$\,0.86 & 1.49\,$\pm$\,1.49 & 0.00\,$\pm$\,0.00 & 0.00\,$\pm$\,0.00 & 1.00\,$\pm$\,0.86 \\
& & CER$\downarrow$ & --- & 52.24\,$\pm$\,1.49 & 63.68\,$\pm$\,3.76 & 55.72\,$\pm$\,3.76 & 47.76\,$\pm$\,1.49 & 71.64\,$\pm$\,2.59 & 68.16\,$\pm$\,4.56 \\
\bottomrule
\end{tabular}
\end{adjustbox}
\caption{Performance of Codex on the \FAB benchmark: mean\,$\pm$\,std over three runs; metrics and notation as in Table~\ref{tab:results_std}.}
\label{tab:results_std_codex}
\end{table*}

\begin{table*}[p]
\centering
\renewcommand{\arraystretch}{1.15}
\setlength{\tabcolsep}{3pt}
\begin{adjustbox}{max width=\textwidth, max totalheight=0.90\textheight}
\begin{tabular}{@{}lll|c|cccccc@{}}
\toprule
\multirow{2}{*}{\rotatebox{90}{\textbf{Task}}} & \multirow{2}{*}{\rotatebox{90}{\textbf{Stock}}} & \multirow{2}{*}{\textbf{Metric}}
& \textbf{BL} & \multicolumn{6}{c}{\textbf{Hermes}} \\
\cmidrule(lr){4-4} \cmidrule(lr){5-10}
& & & B\&H & sonnet & gpt & ds32 & ds4p & qwen9 & qwen4 \\
\midrule
\multirow{12}{*}{\rotatebox{90}{Trading}}
& \multirow{3}{*}{\rotatebox{90}{MSFT}} & CR$\uparrow$ & -23.41 & -1.96\,$\pm$\,3.51 & 3.95\,$\pm$\,4.49 & -7.75\,$\pm$\,3.01 & 1.82\,$\pm$\,6.77 & -1.71\,$\pm$\,3.34 & -16.99\,$\pm$\,7.70 \\
& & SR$\uparrow$ & -3.50 & -0.15\,$\pm$\,0.51 & 0.69\,$\pm$\,0.62 & -1.05\,$\pm$\,0.46 & 0.38\,$\pm$\,0.94 & -0.14\,$\pm$\,0.53 & -3.48\,$\pm$\,1.18 \\
& & MDD$\downarrow$ & 25.12 & 9.89\,$\pm$\,1.35 & 10.27\,$\pm$\,2.09 & 13.37\,$\pm$\,3.01 & 14.60\,$\pm$\,2.29 & 9.58\,$\pm$\,0.40 & 17.40\,$\pm$\,7.55 \\
\cmidrule(lr){2-10}
& \multirow{3}{*}{\rotatebox{90}{TSLA}} & CR$\uparrow$ & -18.80 & -12.97\,$\pm$\,0.52 & 6.57\,$\pm$\,8.45 & \bestup{12.39}\,$\pm$\,9.44 & -5.22\,$\pm$\,2.83 & -6.95\,$\pm$\,10.85 & -4.64\,$\pm$\,4.31 \\
& & SR$\uparrow$ & -2.60 & -1.72\,$\pm$\,0.09 & 0.97\,$\pm$\,1.09 & \bestup{1.88}\,$\pm$\,1.29 & -0.56\,$\pm$\,0.39 & -1.60\,$\pm$\,2.32 & -1.65\,$\pm$\,1.76 \\
& & MDD$\downarrow$ & 21.03 & 19.00\,$\pm$\,0.39 & 13.87\,$\pm$\,2.31 & 9.35\,$\pm$\,2.24 & 20.96\,$\pm$\,2.91 & 13.03\,$\pm$\,3.35 & \bestdown{7.51}\,$\pm$\,1.16 \\
\cmidrule(lr){2-10}
& \multirow{3}{*}{\rotatebox{90}{AAPL}} & CR$\uparrow$ & -6.70 & 3.86\,$\pm$\,1.35 & -5.07\,$\pm$\,6.33 & 1.11\,$\pm$\,6.15 & \bestup{8.82}\,$\pm$\,5.48 & -8.23\,$\pm$\,5.00 & 0.74\,$\pm$\,10.29 \\
& & SR$\uparrow$ & -1.03 & 0.76\,$\pm$\,0.23 & -0.79\,$\pm$\,1.13 & 0.29\,$\pm$\,1.13 & \secondup{1.61}\,$\pm$\,0.96 & -1.61\,$\pm$\,1.08 & 0.28\,$\pm$\,2.40 \\
& & MDD$\downarrow$ & 10.79 & 11.92\,$\pm$\,0.66 & 14.00\,$\pm$\,3.91 & 10.01\,$\pm$\,4.84 & 11.43\,$\pm$\,3.58 & 13.73\,$\pm$\,3.72 & 9.49\,$\pm$\,4.75 \\
\cmidrule(lr){2-10}
& \multirow{3}{*}{\rotatebox{90}{NVDA}} & CR$\uparrow$ & -8.13 & -22.84\,$\pm$\,5.23 & -23.28\,$\pm$\,9.35 & -9.20\,$\pm$\,5.63 & -18.83\,$\pm$\,4.47 & -6.90\,$\pm$\,9.45 & -4.64\,$\pm$\,6.45 \\
& & SR$\uparrow$ & -0.86 & -3.51\,$\pm$\,1.05 & -3.20\,$\pm$\,1.52 & -1.33\,$\pm$\,0.97 & -2.54\,$\pm$\,0.67 & -1.10\,$\pm$\,1.83 & -0.78\,$\pm$\,1.19 \\
& & MDD$\downarrow$ & 14.13 & 23.61\,$\pm$\,5.17 & 25.33\,$\pm$\,7.04 & 16.90\,$\pm$\,1.50 & 20.57\,$\pm$\,3.45 & 11.19\,$\pm$\,6.65 & 11.43\,$\pm$\,4.48 \\
\midrule
\multirow{4}{*}{\rotatebox{90}{Hedging}}
& & PAIR & --- & \shortstack[c]{META\\MSFT\\(3/3)} & \shortstack[c]{MSFT\\NVDA\\(3/3)} & \shortstack[c]{MSFT\\NVDA\\(1/3)} & \shortstack[c]{AMZN\\GOOGL\\(1/3)} & \shortstack[c]{AAPL\\GOOGL\\(2/3)} & \shortstack[c]{NVDA\\TSLA\\(2/3)} \\
& & CR$\uparrow$ & --- & -2.69\,$\pm$\,13.83 & -3.50\,$\pm$\,2.52 & -7.27\,$\pm$\,5.03 & -1.08\,$\pm$\,8.63 & 3.74\,$\pm$\,9.79 & 2.26\,$\pm$\,6.11 \\
& & SR$\uparrow$ & --- & -0.53\,$\pm$\,2.49 & -1.22\,$\pm$\,0.90 & -1.69\,$\pm$\,1.41 & -0.25\,$\pm$\,1.85 & 0.83\,$\pm$\,2.10 & 0.67\,$\pm$\,1.76 \\
& & MDD$\downarrow$ & --- & 10.64\,$\pm$\,7.34 & 6.43\,$\pm$\,1.47 & 11.80\,$\pm$\,1.63 & 9.85\,$\pm$\,2.52 & 7.96\,$\pm$\,3.66 & 4.36\,$\pm$\,1.31 \\
\midrule
\multirow{16}{*}{\rotatebox{90}{Market Insights}}
& \multirow{4}{*}{\rotatebox{90}{MSFT}} & Score$\uparrow$ & --- & 8.47\,$\pm$\,0.04 & 8.39\,$\pm$\,0.15 & 7.56\,$\pm$\,0.50 & 8.60\,$\pm$\,0.01 & 6.96\,$\pm$\,0.64 & 7.22\,$\pm$\,0.40 \\
& & CR$\uparrow$ & -21.74 & -7.21\,$\pm$\,2.61 & -5.30\,$\pm$\,1.57 & -6.57\,$\pm$\,1.62 & -3.20\,$\pm$\,3.05 & -7.41\,$\pm$\,1.25 & -4.65\,$\pm$\,2.90 \\
& & SR$\uparrow$ & -3.25 & -2.57\,$\pm$\,0.99 & -2.19\,$\pm$\,0.66 & -2.13\,$\pm$\,0.64 & \secondup{-1.12}\,$\pm$\,1.11 & -2.53\,$\pm$\,0.52 & -1.74\,$\pm$\,0.95 \\
& & MDD$\downarrow$ & 24.93 & 8.22\,$\pm$\,2.50 & 6.59\,$\pm$\,0.45 & 8.49\,$\pm$\,0.97 & 5.91\,$\pm$\,2.17 & 8.99\,$\pm$\,1.52 & 7.08\,$\pm$\,3.13 \\
\cmidrule(lr){2-10}
& \multirow{4}{*}{\rotatebox{90}{TSLA}} & Score$\uparrow$ & --- & 8.11\,$\pm$\,0.14 & 8.15\,$\pm$\,0.33 & 7.63\,$\pm$\,0.30 & 8.39\,$\pm$\,0.35 & 6.49\,$\pm$\,1.03 & 6.90\,$\pm$\,0.14 \\
& & CR$\uparrow$ & -17.02 & 0.01\,$\pm$\,0.69 & 0.19\,$\pm$\,0.00 & -1.51\,$\pm$\,1.48 & \secondup{1.89}\,$\pm$\,2.94 & 0.45\,$\pm$\,3.60 & 1.36\,$\pm$\,2.35 \\
& & SR$\uparrow$ & -2.35 & 0.15\,$\pm$\,0.57 & 0.13\,$\pm$\,0.00 & -1.40\,$\pm$\,1.49 & 0.59\,$\pm$\,0.79 & 0.31\,$\pm$\,2.99 & 0.49\,$\pm$\,0.85 \\
& & MDD$\downarrow$ & 19.70 & 1.69\,$\pm$\,1.77 & 4.90\,$\pm$\,0.00 & 3.28\,$\pm$\,0.79 & 4.60\,$\pm$\,0.52 & 2.06\,$\pm$\,1.38 & 1.49\,$\pm$\,2.58 \\
\cmidrule(lr){2-10}
& \multirow{4}{*}{\rotatebox{90}{AAPL}} & Score$\uparrow$ & --- & 8.64\,$\pm$\,0.10 & 8.89\,$\pm$\,0.10 & 8.06\,$\pm$\,0.32 & \secondup{9.08}\,$\pm$\,0.12 & 7.04\,$\pm$\,0.65 & 7.36\,$\pm$\,0.39 \\
& & CR$\uparrow$ & -6.14 & 0.02\,$\pm$\,1.83 & 0.04\,$\pm$\,1.07 & -0.49\,$\pm$\,1.27 & 0.43\,$\pm$\,1.18 & -1.66\,$\pm$\,4.17 & -3.72\,$\pm$\,1.67 \\
& & SR$\uparrow$ & -0.94 & 0.03\,$\pm$\,0.94 & 0.08\,$\pm$\,0.47 & -0.05\,$\pm$\,0.33 & 0.23\,$\pm$\,0.54 & -0.78\,$\pm$\,2.04 & -1.15\,$\pm$\,0.52 \\
& & MDD$\downarrow$ & 10.79 & 4.76\,$\pm$\,1.00 & 4.84\,$\pm$\,0.99 & 6.95\,$\pm$\,2.29 & 4.77\,$\pm$\,0.78 & 6.03\,$\pm$\,1.34 & 8.09\,$\pm$\,1.31 \\
\cmidrule(lr){2-10}
& \multirow{4}{*}{\rotatebox{90}{NVDA}} & Score$\uparrow$ & --- & 8.55\,$\pm$\,0.09 & 8.71\,$\pm$\,0.15 & 8.21\,$\pm$\,0.48 & \bestup{9.18}\,$\pm$\,0.03 & 7.52\,$\pm$\,0.38 & 7.83\,$\pm$\,0.17 \\
& & CR$\uparrow$ & -9.05 & -4.04\,$\pm$\,1.98 & -6.50\,$\pm$\,1.17 & -4.67\,$\pm$\,3.13 & -3.45\,$\pm$\,0.19 & -4.85\,$\pm$\,2.79 & -5.46\,$\pm$\,2.26 \\
& & SR$\uparrow$ & -1.00 & -0.96\,$\pm$\,0.53 & -2.09\,$\pm$\,0.56 & -0.94\,$\pm$\,0.52 & -0.86\,$\pm$\,0.03 & -1.30\,$\pm$\,0.79 & -1.33\,$\pm$\,0.77 \\
& & MDD$\downarrow$ & 14.13 & 6.32\,$\pm$\,1.12 & 7.13\,$\pm$\,0.59 & 7.95\,$\pm$\,3.33 & \seconddown{5.06}\,$\pm$\,0.00 & 6.86\,$\pm$\,1.70 & 8.03\,$\pm$\,0.65 \\
\midrule
\multirow{4}{*}{\rotatebox{90}{Auditing}}
& & ACC$\uparrow$ & --- & 43.78\,$\pm$\,3.76 & 32.84\,$\pm$\,2.99 & 37.31\,$\pm$\,2.99 & 51.74\,$\pm$\,0.86 & 30.85\,$\pm$\,0.86 & 29.35\,$\pm$\,4.80 \\
& & SER$\downarrow$ & --- & 0.00\,$\pm$\,0.00 & 7.96\,$\pm$\,3.11 & 15.42\,$\pm$\,1.72 & 0.00\,$\pm$\,0.00 & 5.47\,$\pm$\,0.86 & 3.48\,$\pm$\,2.28 \\
& & EER$\downarrow$ & --- & 2.99\,$\pm$\,0.00 & 1.99\,$\pm$\,0.86 & 1.00\,$\pm$\,1.72 & 0.00\,$\pm$\,0.00 & 0.50\,$\pm$\,0.86 & 0.50\,$\pm$\,0.86 \\
& & CER$\downarrow$ & --- & 53.23\,$\pm$\,3.76 & 57.21\,$\pm$\,5.24 & 46.27\,$\pm$\,2.99 & 48.26\,$\pm$\,0.86 & 63.18\,$\pm$\,0.86 & 66.67\,$\pm$\,1.72 \\
\bottomrule
\end{tabular}
\end{adjustbox}
\caption{Performance of Hermes on the \FAB benchmark: mean\,$\pm$\,std over three runs; metrics and notation as in Table~\ref{tab:results_std}.}
\label{tab:results_std_hermes}
\end{table*}

\begin{table*}[p]
\centering
\renewcommand{\arraystretch}{1.15}
\setlength{\tabcolsep}{3pt}
\begin{adjustbox}{max width=\textwidth, max totalheight=0.90\textheight}
\begin{tabular}{@{}lll|c|cccccc@{}}
\toprule
\multirow{2}{*}{\rotatebox{90}{\textbf{Task}}} & \multirow{2}{*}{\rotatebox{90}{\textbf{Stock}}} & \multirow{2}{*}{\textbf{Metric}}
& \textbf{BL} & \multicolumn{6}{c}{\textbf{OpenClaw}} \\
\cmidrule(lr){4-4} \cmidrule(lr){5-10}
& & & B\&H & sonnet & gpt & ds32 & ds4p & qwen9 & qwen4 \\
\midrule
\multirow{12}{*}{\rotatebox{90}{Trading}}
& \multirow{3}{*}{\rotatebox{90}{MSFT}} & CR$\uparrow$ & -23.41 & 2.82\,$\pm$\,7.71 & \bestup{17.48}\,$\pm$\,8.89 & -15.08\,$\pm$\,11.32 & 7.60\,$\pm$\,11.18 & -7.59\,$\pm$\,4.78 & -12.01\,$\pm$\,3.40 \\
& & SR$\uparrow$ & -3.50 & 0.52\,$\pm$\,1.04 & \bestup{2.65}\,$\pm$\,1.04 & -2.61\,$\pm$\,2.15 & 1.12\,$\pm$\,1.50 & -1.22\,$\pm$\,0.69 & -2.50\,$\pm$\,0.26 \\
& & MDD$\downarrow$ & 25.12 & 10.46\,$\pm$\,1.84 & \bestdown{5.45}\,$\pm$\,0.81 & 19.94\,$\pm$\,4.54 & 11.88\,$\pm$\,2.03 & 12.95\,$\pm$\,1.65 & 14.01\,$\pm$\,5.38 \\
\cmidrule(lr){2-10}
& \multirow{3}{*}{\rotatebox{90}{TSLA}} & CR$\uparrow$ & -18.80 & -11.80\,$\pm$\,3.12 & 3.59\,$\pm$\,1.85 & -2.00\,$\pm$\,5.44 & 5.49\,$\pm$\,12.78 & -11.59\,$\pm$\,6.39 & -9.45\,$\pm$\,4.34 \\
& & SR$\uparrow$ & -2.60 & -1.55\,$\pm$\,0.51 & 0.62\,$\pm$\,0.24 & -0.19\,$\pm$\,0.89 & 0.81\,$\pm$\,1.60 & -2.50\,$\pm$\,1.62 & -2.73\,$\pm$\,0.62 \\
& & MDD$\downarrow$ & 21.03 & 18.98\,$\pm$\,2.03 & 13.89\,$\pm$\,1.55 & 14.60\,$\pm$\,4.17 & 16.01\,$\pm$\,4.05 & 14.46\,$\pm$\,6.33 & 9.99\,$\pm$\,5.23 \\
\cmidrule(lr){2-10}
& \multirow{3}{*}{\rotatebox{90}{AAPL}} & CR$\uparrow$ & -6.70 & -0.71\,$\pm$\,1.71 & 0.01\,$\pm$\,0.88 & -0.36\,$\pm$\,7.59 & 4.38\,$\pm$\,5.07 & -2.90\,$\pm$\,8.49 & -3.81\,$\pm$\,8.49 \\
& & SR$\uparrow$ & -1.03 & -0.01\,$\pm$\,0.30 & 0.12\,$\pm$\,0.15 & 0.04\,$\pm$\,1.37 & 0.89\,$\pm$\,0.89 & -0.56\,$\pm$\,1.81 & -1.29\,$\pm$\,3.01 \\
& & MDD$\downarrow$ & 10.79 & 13.96\,$\pm$\,1.23 & 13.46\,$\pm$\,0.33 & 10.19\,$\pm$\,3.12 & 9.55\,$\pm$\,4.33 & 10.32\,$\pm$\,6.05 & 7.27\,$\pm$\,5.52 \\
\cmidrule(lr){2-10}
& \multirow{3}{*}{\rotatebox{90}{NVDA}} & CR$\uparrow$ & -8.13 & -27.74\,$\pm$\,2.57 & -6.72\,$\pm$\,6.38 & -11.00\,$\pm$\,8.94 & -13.52\,$\pm$\,6.18 & -13.92\,$\pm$\,1.94 & \bestup{2.33}\,$\pm$\,9.02 \\
& & SR$\uparrow$ & -0.86 & -4.29\,$\pm$\,0.52 & -0.73\,$\pm$\,0.84 & -1.46\,$\pm$\,1.32 & -1.64\,$\pm$\,0.86 & -2.29\,$\pm$\,0.26 & \bestup{0.63}\,$\pm$\,1.80 \\
& & MDD$\downarrow$ & 14.13 & 28.46\,$\pm$\,2.55 & 15.06\,$\pm$\,4.13 & 17.28\,$\pm$\,1.83 & 17.09\,$\pm$\,3.29 & 16.60\,$\pm$\,1.56 & \bestdown{7.11}\,$\pm$\,2.66 \\
\midrule
\multirow{4}{*}{\rotatebox{90}{Hedging}}
& & PAIR & --- & \shortstack[c]{META\\MSFT\\(2/3)} & \shortstack[c]{MSFT\\NVDA\\(2/3)} & \shortstack[c]{NVDA\\MSFT\\(2/3)} & \shortstack[c]{NVDA\\MSFT\\(1/3)} & \shortstack[c]{META\\MSFT\\(1/3)} & \shortstack[c]{AAPL\\META\\(1/3)} \\
& & CR$\uparrow$ & --- & -5.14\,$\pm$\,9.53 & -1.18\,$\pm$\,4.69 & -6.29\,$\pm$\,4.74 & -4.22\,$\pm$\,6.57 & -3.53\,$\pm$\,2.87 & \secondup{6.10}\,$\pm$\,4.41 \\
& & SR$\uparrow$ & --- & -1.21\,$\pm$\,2.06 & -0.40\,$\pm$\,1.68 & -1.41\,$\pm$\,1.20 & -0.82\,$\pm$\,1.43 & -0.74\,$\pm$\,0.71 & \secondup{1.42}\,$\pm$\,1.01 \\
& & MDD$\downarrow$ & --- & 11.30\,$\pm$\,4.81 & 7.27\,$\pm$\,3.68 & 11.23\,$\pm$\,1.81 & 10.24\,$\pm$\,2.71 & 8.90\,$\pm$\,1.44 & 5.84\,$\pm$\,1.53 \\
\midrule
\multirow{16}{*}{\rotatebox{90}{Market Insights}}
& \multirow{4}{*}{\rotatebox{90}{MSFT}} & Score$\uparrow$ & --- & 8.20\,$\pm$\,0.14 & 8.33\,$\pm$\,0.08 & 5.92\,$\pm$\,1.27 & \secondup{8.61}\,$\pm$\,0.14 & 5.46\,$\pm$\,0.41 & 6.55\,$\pm$\,0.68 \\
& & CR$\uparrow$ & -21.74 & -6.62\,$\pm$\,0.98 & -3.98\,$\pm$\,3.18 & -7.74\,$\pm$\,4.73 & -6.96\,$\pm$\,2.16 & -7.12\,$\pm$\,0.59 & -5.30\,$\pm$\,2.65 \\
& & SR$\uparrow$ & -3.25 & -2.23\,$\pm$\,0.45 & -1.82\,$\pm$\,1.00 & -2.69\,$\pm$\,1.37 & -2.32\,$\pm$\,0.69 & -2.31\,$\pm$\,0.33 & -2.56\,$\pm$\,1.26 \\
& & MDD$\downarrow$ & 24.93 & 8.73\,$\pm$\,1.31 & 5.11\,$\pm$\,3.22 & 9.41\,$\pm$\,4.68 & 9.00\,$\pm$\,1.37 & 8.67\,$\pm$\,1.26 & 6.98\,$\pm$\,2.63 \\
\cmidrule(lr){2-10}
& \multirow{4}{*}{\rotatebox{90}{TSLA}} & Score$\uparrow$ & --- & 8.21\,$\pm$\,0.05 & 8.31\,$\pm$\,0.09 & 5.83\,$\pm$\,0.61 & 8.38\,$\pm$\,0.08 & 6.36\,$\pm$\,0.53 & 5.98\,$\pm$\,0.55 \\
& & CR$\uparrow$ & -17.02 & -0.26\,$\pm$\,0.49 & 0.36\,$\pm$\,0.33 & -0.79\,$\pm$\,2.17 & 1.35\,$\pm$\,2.82 & -0.70\,$\pm$\,1.99 & 0.00\,$\pm$\,0.00 \\
& & SR$\uparrow$ & -2.35 & -0.34\,$\pm$\,0.65 & 0.24\,$\pm$\,0.19 & -0.78\,$\pm$\,2.28 & 0.48\,$\pm$\,0.99 & -0.85\,$\pm$\,1.93 & 0.00\,$\pm$\,0.00 \\
& & MDD$\downarrow$ & 19.70 & \seconddown{1.22}\,$\pm$\,1.13 & 3.88\,$\pm$\,0.96 & 2.01\,$\pm$\,1.12 & 4.30\,$\pm$\,1.07 & 2.20\,$\pm$\,1.91 & 0.00\,$\pm$\,0.00 \\
\cmidrule(lr){2-10}
& \multirow{4}{*}{\rotatebox{90}{AAPL}} & Score$\uparrow$ & --- & 8.84\,$\pm$\,0.03 & 8.63\,$\pm$\,0.07 & 7.03\,$\pm$\,0.80 & \bestup{9.17}\,$\pm$\,0.14 & 6.93\,$\pm$\,0.36 & 6.83\,$\pm$\,0.57 \\
& & CR$\uparrow$ & -6.14 & -1.38\,$\pm$\,0.21 & -0.73\,$\pm$\,0.75 & -2.68\,$\pm$\,2.87 & 0.55\,$\pm$\,0.91 & -1.76\,$\pm$\,3.85 & \bestup{1.47}\,$\pm$\,0.72 \\
& & SR$\uparrow$ & -0.94 & -0.48\,$\pm$\,0.11 & -0.36\,$\pm$\,0.40 & -0.88\,$\pm$\,0.96 & 0.30\,$\pm$\,0.38 & -0.69\,$\pm$\,1.76 & \secondup{0.93}\,$\pm$\,0.80 \\
& & MDD$\downarrow$ & 10.79 & 6.13\,$\pm$\,0.32 & 4.51\,$\pm$\,0.29 & 6.50\,$\pm$\,2.99 & 4.31\,$\pm$\,0.52 & 5.18\,$\pm$\,1.32 & 4.62\,$\pm$\,3.53 \\
\cmidrule(lr){2-10}
& \multirow{4}{*}{\rotatebox{90}{NVDA}} & Score$\uparrow$ & --- & 8.75\,$\pm$\,0.05 & 8.48\,$\pm$\,0.16 & 5.93\,$\pm$\,0.57 & \secondup{9.11}\,$\pm$\,0.11 & 6.77\,$\pm$\,0.39 & 6.66\,$\pm$\,0.90 \\
& & CR$\uparrow$ & -9.05 & -4.47\,$\pm$\,0.29 & -6.38\,$\pm$\,0.00 & -7.52\,$\pm$\,0.96 & -5.25\,$\pm$\,2.32 & -2.47\,$\pm$\,3.06 & -7.26\,$\pm$\,5.14 \\
& & SR$\uparrow$ & -1.00 & -1.05\,$\pm$\,0.09 & -2.05\,$\pm$\,0.00 & -1.77\,$\pm$\,0.41 & -1.11\,$\pm$\,0.30 & -0.61\,$\pm$\,0.89 & -1.61\,$\pm$\,1.14 \\
& & MDD$\downarrow$ & 14.13 & 7.37\,$\pm$\,0.28 & 7.12\,$\pm$\,0.00 & 10.51\,$\pm$\,0.95 & 7.96\,$\pm$\,3.34 & 6.20\,$\pm$\,1.32 & 10.02\,$\pm$\,4.97 \\
\midrule
\multirow{4}{*}{\rotatebox{90}{Auditing}}
& & ACC$\uparrow$ & --- & 48.76\,$\pm$\,3.11 & 30.85\,$\pm$\,1.72 & 46.77\,$\pm$\,2.28 & \secondup{52.74}\,$\pm$\,0.86 & 25.37\,$\pm$\,0.00 & 28.86\,$\pm$\,2.28 \\
& & SER$\downarrow$ & --- & 0.50\,$\pm$\,0.86 & 0.00\,$\pm$\,0.00 & 1.49\,$\pm$\,2.59 & 1.00\,$\pm$\,0.86 & 7.96\,$\pm$\,6.03 & 6.47\,$\pm$\,2.28 \\
& & EER$\downarrow$ & --- & 1.99\,$\pm$\,0.86 & 2.99\,$\pm$\,0.00 & 1.00\,$\pm$\,0.86 & 0.00\,$\pm$\,0.00 & 0.00\,$\pm$\,0.00 & 0.00\,$\pm$\,0.00 \\
& & CER$\downarrow$ & --- & 48.76\,$\pm$\,3.45 & 66.17\,$\pm$\,1.72 & 50.75\,$\pm$\,2.59 & 46.27\,$\pm$\,0.00 & 66.67\,$\pm$\,6.03 & 64.68\,$\pm$\,3.45 \\
\bottomrule
\end{tabular}
\end{adjustbox}
\caption{Performance of OpenClaw on the \FAB benchmark: mean\,$\pm$\,std over three runs; metrics and notation as in Table~\ref{tab:results_std}.}
\label{tab:results_std_openclaw}
\end{table*}

\section{Live Benchmark}
\label{sec:sec_live}

We additionally evaluate the three market-based workflows on a post-release window from August 3 to September 30, 2026 (42 trading days), which begins after every evaluated backbone model was publicly released (Table~\ref{tab:agent_release_dates}); its outcomes therefore cannot be contained in any backbone's parameters. Before each interaction, the environment incorporates records newly available by the corresponding decision date. The live setting preserves the same asset universes, peer mappings, tool interfaces, interaction protocols, and realized-outcome metrics as the offline evaluation; Market Insights produces nine weekly reports per asset, from August 7 to September 30. For each workflow, we evaluate three agent--backbone configurations selected on the offline window, with the selection fixed before any live run. As in the offline evaluation (Appendix~\ref{app:multirun}), each configuration is run three times independently, metrics are computed within each run and then aggregated, and decision steps without a committed submission are retained as failures rather than excluded. Table~\ref{tab:live_std} reports the mean together with the sample standard deviation.

% generated by analysis/build_live_table.py -- do not edit by hand
% needs table/tab_macros.tex
\begin{table*}[!tp]
\centering
\footnotesize
\renewcommand{\arraystretch}{1.0}
\setlength{\tabcolsep}{6pt}
\begin{tabular}{lll|c|ccc}
\toprule
\textbf{Task} & \textbf{Stock} & \textbf{Metric} & \textbf{BL} & \multicolumn{3}{c}{\textbf{Selected configurations}} \\
\midrule
& & & \textit{B\&H} & \makecell{\textit{Claude Code}\\\texttt{gpt}} & \makecell{\textit{OpenClaw}\\\texttt{gpt}} & \makecell{\textit{OpenClaw}\\\texttt{ds4p}} \\
\cmidrule(lr){4-7}
\multirow{12}{*}{Trading}
& \multirow{3}{*}{MSFT} & CR$\uparrow$ & 7.93 & \bestup{-1.74}\,$\pm$\,1.15 & -2.19\,$\pm$\,3.13 & -10.37\,$\pm$\,1.63 \\
& & SR$\uparrow$ & 2.24 & \bestup{-0.42}\,$\pm$\,0.34 & -0.56\,$\pm$\,0.96 & -3.07\,$\pm$\,0.57 \\
& & MDD$\downarrow$ & 4.81 & \bestdown{9.02}\,$\pm$\,1.06 & 10.00\,$\pm$\,1.74 & 13.54\,$\pm$\,0.33 \\
\cmidrule(lr){2-7}
& \multirow{3}{*}{TSLA} & CR$\uparrow$ & 14.10 & -8.11\,$\pm$\,7.60 & \bestup{-6.86}\,$\pm$\,8.31 & -15.17\,$\pm$\,7.02 \\
& & SR$\uparrow$ & 2.49 & -1.59\,$\pm$\,1.67 & \bestup{-1.15}\,$\pm$\,1.60 & -3.16\,$\pm$\,1.68 \\
& & MDD$\downarrow$ & 8.45 & \bestdown{15.10}\,$\pm$\,2.55 & 15.83\,$\pm$\,5.67 & 22.52\,$\pm$\,1.86 \\
\cmidrule(lr){2-7}
& \multirow{3}{*}{AAPL} & CR$\uparrow$ & 7.66 & 0.26\,$\pm$\,7.04 & \bestup{1.99}\,$\pm$\,7.68 & -2.44\,$\pm$\,2.74 \\
& & SR$\uparrow$ & 2.06 & 0.13\,$\pm$\,1.88 & \bestup{0.58}\,$\pm$\,2.03 & -0.60\,$\pm$\,0.83 \\
& & MDD$\downarrow$ & 3.90 & \bestdown{5.36}\,$\pm$\,1.91 & 6.13\,$\pm$\,3.57 & 8.40\,$\pm$\,0.74 \\
\cmidrule(lr){2-7}
& \multirow{3}{*}{NVDA} & CR$\uparrow$ & 15.65 & \bestup{12.39}\,$\pm$\,6.48 & 7.55\,$\pm$\,14.81 & 4.87\,$\pm$\,6.12 \\
& & SR$\uparrow$ & 2.70 & \bestup{2.19}\,$\pm$\,1.02 & 1.32\,$\pm$\,2.51 & 0.98\,$\pm$\,1.02 \\
& & MDD$\downarrow$ & 9.28 & \bestdown{9.02}\,$\pm$\,2.94 & 9.59\,$\pm$\,3.67 & 10.82\,$\pm$\,4.49 \\
\midrule
& & & \textit{B\&H} & \makecell{\textit{Claude Code}\\\texttt{qwen9}} & \makecell{\textit{Claude Code}\\\texttt{qwen4}} & \makecell{\textit{OpenClaw}\\\texttt{qwen4}} \\
\cmidrule(lr){4-7}
\multirow{4}{*}{Hedging}
& & PAIR & --- & \shortstack[c]{AMZN\\GOOGL\\(1/3)} & \shortstack[c]{NVDA\\TSLA\\(1/3)} & \shortstack[c]{GOOGL\\META\\(1/3)} \\
& & CR$\uparrow$ & --- & 0.37\,$\pm$\,2.58 & \bestup{2.22}\,$\pm$\,6.86 & -3.61\,$\pm$\,5.42 \\
& & SR$\uparrow$ & --- & 0.44\,$\pm$\,1.21 & \bestup{0.50}\,$\pm$\,2.22 & -0.87\,$\pm$\,2.08 \\
& & MDD$\downarrow$ & --- & 5.73\,$\pm$\,2.97 & \bestdown{5.31}\,$\pm$\,0.77 & 7.84\,$\pm$\,4.56 \\
\midrule
& & & \textit{B\&H} & \makecell{\textit{Claude Code}\\\texttt{ds32}} & \makecell{\textit{Claude Code}\\\texttt{qwen4}} & \makecell{\textit{Hermes}\\\texttt{ds4p}} \\
\cmidrule(lr){4-7}
\multirow{16}{*}{\makecell[l]{Market\\Insights}}
& \multirow{4}{*}{MSFT} & Score$\uparrow$ & --- & 5.09\,$\pm$\,1.00 & 6.22\,$\pm$\,0.06 & \bestup{8.82}\,$\pm$\,0.11 \\
& & CR$\uparrow$ & 2.07 & -1.72\,$\pm$\,1.27 & -0.71\,$\pm$\,1.81 & \bestup{0.20}\,$\pm$\,2.17 \\
& & SR$\uparrow$ & 0.80 & -2.03\,$\pm$\,1.28 & -0.29\,$\pm$\,0.99 & \bestup{0.02}\,$\pm$\,1.78 \\
& & MDD$\downarrow$ & 4.81 & \bestdown{2.67}\,$\pm$\,1.21 & 2.76\,$\pm$\,0.58 & 2.71\,$\pm$\,0.73 \\
\cmidrule(lr){2-7}
& \multirow{4}{*}{TSLA} & Score$\uparrow$ & --- & 4.15\,$\pm$\,1.99 & 6.41\,$\pm$\,0.03 & \bestup{8.63}\,$\pm$\,0.13 \\
& & CR$\uparrow$ & 8.64 & \bestup{3.14}\,$\pm$\,2.72 & -0.51\,$\pm$\,1.84 & 2.96\,$\pm$\,3.02 \\
& & SR$\uparrow$ & 1.82 & \bestup{2.26}\,$\pm$\,2.19 & -0.10\,$\pm$\,0.91 & 1.48\,$\pm$\,1.58 \\
& & MDD$\downarrow$ & 8.45 & \bestdown{1.11}\,$\pm$\,1.12 & 2.24\,$\pm$\,2.23 & 2.24\,$\pm$\,0.00 \\
\cmidrule(lr){2-7}
& \multirow{4}{*}{AAPL} & Score$\uparrow$ & --- & 4.12\,$\pm$\,1.49 & 6.05\,$\pm$\,0.46 & \bestup{8.62}\,$\pm$\,0.11 \\
& & CR$\uparrow$ & 8.54 & 0.15\,$\pm$\,0.75 & \bestup{0.34}\,$\pm$\,0.82 & -1.26\,$\pm$\,1.09 \\
& & SR$\uparrow$ & 2.65 & \bestup{0.50}\,$\pm$\,1.14 & 0.22\,$\pm$\,0.66 & -1.26\,$\pm$\,1.01 \\
& & MDD$\downarrow$ & 3.90 & 1.83\,$\pm$\,0.94 & \bestdown{1.16}\,$\pm$\,1.03 & 2.06\,$\pm$\,0.70 \\
\cmidrule(lr){2-7}
& \multirow{4}{*}{NVDA} & Score$\uparrow$ & --- & 2.98\,$\pm$\,1.02 & 6.71\,$\pm$\,0.18 & \bestup{8.78}\,$\pm$\,0.37 \\
& & CR$\uparrow$ & 2.34 & \bestup{-0.68}\,$\pm$\,7.23 & -3.68\,$\pm$\,2.57 & -1.29\,$\pm$\,2.90 \\
& & SR$\uparrow$ & 0.66 & \bestup{0.43}\,$\pm$\,3.82 & -2.08\,$\pm$\,1.55 & -0.59\,$\pm$\,1.40 \\
& & MDD$\downarrow$ & 9.28 & \bestdown{4.20}\,$\pm$\,4.54 & 6.16\,$\pm$\,1.31 & 5.42\,$\pm$\,1.08 \\
\bottomrule
\end{tabular}
\caption{
Live performance of frontier AI agents on the three market-based workflows of \FAB (post-release window, 2026-08-03 to 2026-09-30).
For each workflow, the three configurations with the highest mean offline Sharpe ratio (Table~\ref{tab:results}) are evaluated; the selection was fixed before any live run.
Each cell reports mean\,$\pm$\,std over three independent runs; metrics are computed per run, then aggregated.
\textbf{Metrics.} As in Table~\ref{tab:results}; Market Insights positions start on the first trading day after the first report (2026-08-10), and its B\&H baseline covers the same window.
\textbf{Notation.} B\&H: Buy\&Hold baseline (BL). The best of the three configurations is shaded in bold (MDD is ranked only among configurations that held a position); blue: higher is better, purple: lower is better. ``---'': not applicable.
}
\label{tab:live_std}
\end{table*}

\clearpage
\section{Raw Data Source}
\label{sec:sec_raw}
This appendix documents the source systems, collection periods, and
processing steps for each data modality used in \FAB. The underlying
financial records are publicly available and contain no personally
identifiable information.

\paragraph{Availability and licenses}
We will publicly release the \FAB environments, task instances, evaluation
code, and the per-step run records behind all reported results upon
publication; an anonymized version is provided as supplementary material
for review. The code is released under the MIT License, and the data we create (task
instances, rubrics, LLM-generated news summaries, and run records) under
CC BY 4.0. Upstream sources are used under their own terms. SEC EDGAR
filings and XBRL packages are public U.S.\ government records, and the
US-GAAP Taxonomy is used under the terms of the FASB. Raw prices (Yahoo
Finance via \texttt{yfinance}) and news articles (OpenAI Web Search,
Finnhub, NewsData, \texttt{yfinance}, and Google News RSS for the
post-release window) are not redistributed publicly: the public release
provides the scripts that retrieve them, together with the news
summaries used by the environments (Appendix~\ref{sec:sec_news}). The
price tables included in the review supplementary material are provided
to reviewers only, for reproducibility.

\subsection{Stock Price}
\label{sec:sec_price}

We collect daily OHLCV data (open, high, low, close, and volume) and
adjusted close prices for the eight equities in our universe:
\texttt{AAPL}, \texttt{ADBE}, \texttt{AMZN}, \texttt{GOOGL},
\texttt{META}, \texttt{MSFT}, \texttt{NVDA}, and \texttt{TSLA}.
The offline collection spans December 1, 2024 to March 31, 2026,
providing approximately thirteen months of queryable history before
the three-month evaluation horizon. Live evaluation runs from
August 3 to September 30, 2026, with newly available price records appended
before each daily interaction.

Prices are retrieved from Yahoo Finance through the open-source
\texttt{yfinance}~\citep{yfinance}
Python library. We use adjusted close when computing strategy returns,
as it accounts for stock splits and cash dividends and provides a
continuous return series. Yahoo Finance reports only U.S.\ trading
days; weekends, exchange holidays, and unscheduled closures are
therefore absent, and we perform no imputation or forward-filling.

\subsection{Corporate Filings (10-K / 10-Q)}
\label{sec:sec_filings}

We retrieve 10-K and 10-Q filings for each equity through \texttt{SEC API}~\citep{secapi}, a commercial interface around SEC EDGAR. Backtesting uses filings with \texttt{filed\_at} between 2024-01-01 and 2026-03-31, and the live evaluation covers 2026-08-03 to 2026-09-30, with newly filed
documents are incorporated before the corresponding interaction.

We filter by form type (\texttt{10-K} or \texttt{10-Q}) and ticker, and for every filing returned we extract two sections as plain text: Management's Discussion \& Analysis (Item~7 in 10-K; Part~I Item~2 in 10-Q) and Risk Factors (Item~1A in 10-K; Part~II Item~1A in 10-Q). HTML and iXBRL parsing is handled by the API.

Two ticker and form-handling decisions are worth noting. Alphabet files
its reports under the SEC ticker \texttt{GOOG}; we therefore query
\texttt{GOOG} and relabel the resulting records as \texttt{GOOGL} to
match the canonical symbol used throughout \FAB. Amendment forms
(10-K/A and 10-Q/A) are excluded because they are not supported by the
section extractor; this removes two amended TSLA filings.

To prevent look-ahead leakage, each filing becomes accessible according
to its \texttt{filed\_at} timestamp, rather than its
\texttt{period\_of\_report}. This distinction is material because a
filing describing an earlier fiscal period is typically released weeks
later. For example, TSLA's Q1-2026 10-Q reports a fiscal period ending
March 31, 2026 but was filed on April 22, 2026; indexing it by the
reporting period would make the document available before its public
release.

\subsection{News Summarization}
\label{sec:sec_news}

Following prior work
~\citep{qian2025agentstradelivemultimarket}, we collect asset-related
news from multiple sources, including OpenAI Web
Search~\citep{openai_websearch},
Finnhub~\citep{finnhub},
NewsData~\citep{newsdata}, and
\texttt{yfinance}~\citep{yfinance}.
Articles are retrieved for each asset and date using its ticker and
company name, and overlapping reports of the same event are
consolidated.

We use GPT-5-nano to convert the retained articles into daily
asset-level news records. The summarization prompt requires the model
to rely only on the retrieved articles, introduce no external
information or price forecasts, and focus on reported events,
sentiment, and relevant market context. Each record is indexed by its
publication date and becomes accessible only from that date onward.

\subsection{XBRL Filings}
\label{sec:sec_xbrl}

Following the collection procedure of
FinAuditing~\citep{wang2026finauditingfinancialtaxonomystructuredmultidocument},
we identify SEC filings associated with validation messages released
by the XBRL US Data Quality Committee
(DQC)~\citep{xbrl_dqc}
and retrieve the corresponding filing packages from SEC
EDGAR~\citep{sec_edgar}. The filings are
collected independently for \FAB and do not reuse instances from the
FinAuditing benchmark.

For each retained filing, we preserve the XBRL instance document,
schema, calculation linkbase, definition linkbase, label linkbase, and
presentation linkbase. We normalize filing metadata, remove incomplete
or inaccessible packages, and align each filing with its corresponding
DQC validation information. The resulting collection is used to
construct the 67 Auditing instances described in
Section~\ref{sec:sec_auditing}. DQC messages guide instance selection
but are not exposed to agents during evaluation.

\subsection{US-GAAP Taxonomy}
\label{sec:sec_gaap}

Following FinAuditing
~\citep{wang2026finauditingfinancialtaxonomystructuredmultidocument},
we collect the official US-GAAP Taxonomy releases from the Financial
Accounting Standards Board (FASB) for 2020--2024. Each filing is paired
with the taxonomy release referenced by its XBRL package.

For each release, we parse the schema and associated linkbases into a
year-specific taxonomy index. The schema provides concept attributes,
including name, data type, period type, substitution group, namespace,
and balance type. The label linkbase provides human-readable and
documentation labels, while the calculation linkbase records numerical
parent--child relations and their weights. We additionally retain
presentation and definition relations when available.

These indices support the \texttt{query\_taxonomy} tools used in the
Auditing environment, including retrieval of concept definitions,
labels, metadata, and calculation relations. Taxonomy releases remain
separate by year; shared concept names are exposed through a consistent
query interface without merging their year-specific definitions or
relationships.

\section{Agent Implementation Details}
\label{app:implement}

This appendix describes the five agent frameworks evaluated in \FAB.
For frameworks that expose a configurable reasoning level, we set it to
\texttt{medium}. Persistent cross-run memory and built-in web search are
disabled so that agents operate only on information available through
the benchmark environment.

Every framework connects to the same MCP server for each workflow,
which serves the instructions and assignment (\texttt{get\_task};
Appendix~\ref{sec:sec_prompts}), the workflow tools, and a write-once
\texttt{submit} tool. Information-access constraints are enforced by
the server, and a step counts as completed only if a submission was
accepted. Each step is limited to 25 turns, 50 tool calls (100 in
Hedging), 8{,}000 output tokens per response, and 600 seconds; a step
that ends without a submission is retained as a missing decision.
The frameworks otherwise retain their native planning, state
management, and tool-orchestration policies.

\begin{table}[h]
\centering
\small
\setlength{\tabcolsep}{5pt}
\begin{tabular}{l l l}
\toprule
\textbf{Component} & \textbf{Name} & \textbf{Release Date} \\
\midrule
\multirow{6}{*}{Backbone Model}
 & Claude Sonnet 5        & Jun 30, 2026 \\
 & GPT-5.4                & Mar 5, 2026 \\
 & DeepSeek-V3.2          & Dec 1, 2025 \\
 & DeepSeek-V4-Pro        & Apr 24, 2026$^{b}$ \\
 & Qwen3.5-9B             & Mar 2, 2026 \\
 & Qwen3.5-4B             & Mar 2, 2026 \\
\midrule
\multirow{5}{*}{Agent Framework}
 & ReAct                  & Oct 6, 2022$^{a}$ \\
 & Claude Code            & Feb 24, 2025 \\
 & Codex CLI              & Apr 16, 2025 \\
 & Hermes Agent           & Feb 25, 2026 \\
 & OpenClaw               & Nov 24, 2025$^{c}$ \\
\bottomrule
\end{tabular}

\vspace{2pt}
\begin{minipage}{\columnwidth}
\small
$^{a}$Date of the original ReAct paper rather than a software release.
$^{b}$Preview release, used before Aug 13, 2026; the general-availability version, released on Aug 13, 2026 under the same API model name, is used from that date on.
$^{c}$First released as Clawdbot; renamed OpenClaw on Jan 30, 2026.
\end{minipage}

\caption{
Public release dates of the evaluated backbone models and agent
frameworks. All backbone models were first released before the
post-release live evaluation (Aug 3--Sep 30, 2026).
}
\label{tab:agent_release_dates}
\end{table}

\paragraph{ReAct Agent}
We implement a minimal ReAct-style reasoning--action loop
~\citep{yao2022react} using
LangChain~\citep{langchain}. At each step,
the backbone receives the accumulated interaction trajectory and
available tools, then either invokes a tool or proceeds toward the
required workflow output. The controller executes each tool call and
returns the resulting observation to the backbone.

We expose the workflow-specific MCP server as the agent's tool set. No
additional planner, sub-agent, or external memory is used. ReAct
therefore serves as the minimal reference framework in our
agent--backbone comparison.

\paragraph{Claude Code}
Claude Code is an execution-oriented agent with native shell,
filesystem, and MCP support~\citep{claudecode}.
We run it in headless mode and register each workflow-specific MCP
server through its \texttt{.mcp.json} configuration.

For each interaction, Claude Code obtains its instructions and submits
its output through the MCP tools.
Tools unrelated to \FAB are not enabled. This configuration retains
Claude Code's native planning and tool-orchestration behavior while
restricting its accessible information and operations to the benchmark
environment.

\paragraph{Codex}
Codex is an execution-oriented agent with native shell, filesystem, and
MCP support~\citep{codex}. We run it
in headless mode and register each workflow-specific MCP server through its
configuration.

Codex receives the same workflow instructions and external tool surface
as the other frameworks, while retaining its native planning and
tool-orchestration policy. Tools unrelated to \FAB are disabled.

\paragraph{Hermes}
Hermes is an open-source general-purpose agent with a unified provider
interface and native MCP support~\citep{hermesagent}.
We run Hermes through its CLI, configure each backbone as the primary
model provider, and register the workflow-specific MCP server as its
external tool source.

Autonomous skill creation and cross-session memory are disabled.
Hermes otherwise retains its native interaction loop and
tool-orchestration policy while operating within the same workflow
instructions and information boundaries as the other frameworks.

\paragraph{OpenClaw}
OpenClaw is an open-source agent framework with a plugin-based tool
interface~\citep{openclaw}. We run it locally,
and register each workflow-specific MCP server through its tool-plugin
interface.

Browser, messaging, and other tools unrelated to \FAB are disabled.
OpenClaw otherwise retains its native reasoning and tool-execution
policies while operating under the same workflow instructions and
information boundaries as the other frameworks.

\paragraph{Execution settings}
All runs use the same limits per decision step: at most \(25\) model
turns, \(8{,}000\) output tokens per model call, a \(600\)-second
timeout, a workflow-specific cap on tool calls (\(50\) for Trading,
Market Insights, and Auditing; \(100\) for Hedging, whose pair selection
alone can require \(28\) pair comparisons), and a cap on the tokens
processed in a step (\(120\)k for Trading, \(200\)k for Hedging,
\(250\)k for Market Insights, \(150\)k for Auditing). A step that fails
because of the infrastructure or times out is attempted at most twice;
reaching a limit or ending without a submission is final and scored as
such. Hedging steps of an earlier pilot with a cap of \(50\) tool calls
were rerun with the final cap. Reasoning effort is set to
\texttt{medium} for every backbone, and sampling uses the provider
defaults. API backbones are accessed through an OpenAI-compatible
gateway; Qwen3.5-9B and Qwen3.5-4B are served with vLLM on one NVIDIA
A800 GPU each, with their native \(262\)k-token context. Harness versions
are the Claude Agent SDK 0.2.165 (Claude Code), Codex CLI 0.162.1, Hermes
Agent 0.19.0, and OpenClaw 2026.9.9. Claude Code reaches non-Claude
backbones through a LiteLLM translation layer, and Codex reaches the
DeepSeek and Qwen backbones through a Chat Completions bridge, because
their Responses endpoints do not accept the reasoning-effort setting.

\section{Experimental Prompts}
\label{sec:sec_prompts}

We provide the task instructions for each workflow and the complete
prompt used for rubric generation. The instructions are identical for
every agent, backbone, and run; the agent obtains them, together with
its instance-specific assignment (asset, decision or report date, and
information cut-off), from the \texttt{get\_task} tool of the workflow's
MCP server and commits its decision through \texttt{submit}. All
instructions and server code are released with the code.

\begin{promptlisting}{Trading Instructions}
# Trading task

You manage a position in one U.S. stock over a sequence of trading days. Each
step of the task is one trading day. Your assignment from `get_task` gives the
stock, the decision date, and the date through which information is available.

## Timing
- You decide **before the market opens** on the decision date, using
  information through the close of the previous trading day.
- Your decision takes effect at the **open** of the decision date and is held
  until the open of the next trading day, when you decide again.
- Tools never return information dated after the information cut-off.

## Actions
Submit exactly one action per step with `submit`:
- `BUY`: hold a long position (+1x the stock) for the day.
- `SELL`: hold a short position (-1x) for the day.
- `HOLD`: hold no position for the day.

There are no transaction costs. Your performance is the realised return of the
positions you choose over the whole sequence.

## Tools
| Tool | Use |
|---|---|
| `get_task()` | Assignment, current position, instructions. |
| `get_history(last_n)` | Your previous decisions in this task and their realised returns so far. |
| `get_prices(symbol, date_start, date_end)` | Daily open/high/low/close/volume. |
| `get_indicator(symbol, indicator, date_start, date_end, length?)` | `ma`, `rsi`, `bbands` or `macd`. |
| `list_news(symbol, date_start, date_end)` / `get_news(symbol, news_id)` | Daily news digests. |
| `list_filings(symbol, date_start, date_end)` / `get_filing_section(symbol, filing_id, section)` | 10-K/10-Q MD&A (`mda`) and Risk Factors (`risk`). |
| `submit(action)` | Your decision for this step. Accepted exactly once. |

Your task step is complete only after `submit` returns `"accepted": true`.
\end{promptlisting}

\begin{promptlisting}{Hedging Instructions}
# Hedging task

You run a market-neutral pair position over a sequence of trading days. Each
step is one trading day. The assignment from `get_task` gives the stock
universe, the decision date, the information cut-off and your current state.

## Pair selection
On your first decision you choose an ordered pair `[A, B]` of two different
stocks from the universe. The pair is then fixed for the rest of the task.

## Timing
- You decide **before the market opens** on the decision date, using
  information through the close of the previous trading day.
- Positions change at the **open** of the decision date and are held until
  the open of the next trading day.

## Actions (for the pair `[A, B]`)
- `LONG_SHORT`: long A and short B, each with half of the capital (+0.5 A, -0.5 B).
- `SHORT_LONG`: short A and long B (-0.5 A, +0.5 B).
- `HOLD`: keep the current position (or stay flat if you have none).
- `CLOSE`: exit any position and stay flat.

There are no transaction costs. Your performance is the realised return of the
pair positions over the whole sequence.

## Tools
| Tool | Use |
|---|---|
| `get_task()` | Assignment, locked pair (if any), current position. |
| `get_history(last_n)` | Your previous decisions and their realised returns so far. |
| `compare_pair(symbol_a, symbol_b, lookback_days?)` | Aligned prices, return correlation, hedge ratio and spread statistics for two stocks. |
| `get_prices`, `get_indicator`, `list_news`/`get_news`, `list_filings`/`get_filing_section` | Per-stock data, as in their descriptions. |
| `submit(action, pair?)` | Your decision. Include `pair` only on the decision that selects it. Accepted exactly once per step. |

Your task step is complete only after `submit` returns `"accepted": true`.
\end{promptlisting}

\begin{promptlisting}{Market Insights Instructions}
# Market Insights task

You write a weekly equity research report on one U.S. stock and commit to an
investment rating. Each step is one report week. The assignment from `get_task`
gives the stock, the report date (the last trading day of the week) and the
information cut-off (the close of the report date).

## The report
Markdown with these eight sections, as level-2 headings, in this order:
1. `## Executive Summary`
2. `## Investment Thesis and Rating`
3. `## Price Performance and Technicals`
4. `## News and Catalysts`
5. `## Earnings and Filings`
6. `## Sector-Relative Performance`
7. `## Risks`
8. `## Outlook and Scenarios`

Use only information available through the cut-off and say where each claim
comes from (prices, the weekly statistics, a dated news digest, or a filing).

## The rating
One of `STRONG_BUY`, `BUY`, `HOLD`, `SELL`, `STRONG_SELL`. The rating takes
effect at the next market open and is held until the next report's rating takes
effect. It is evaluated as a position in the stock: `STRONG_BUY` +1x, `BUY`
+0.5x, `HOLD` 0, `SELL` -0.5x, `STRONG_SELL` -1x. The report is also evaluated
for accuracy, fidelity to the evidence and consistency with the rating.

## Tools
| Tool | Use |
|---|---|
| `get_task()` | Assignment, your previous rating, instructions. |
| `get_history(last_n)` | Your previous ratings and their realised returns so far. |
| `get_weekly_metrics()` | 16 deterministic statistics for the week (performance, momentum, sector-relative). |
| `get_prices`, `get_indicator`, `list_news`/`get_news`, `list_filings`/`get_filing_section` | Data for the stock and its peers. |
| `submit(rating, report_markdown)` | Your report and rating. Accepted exactly once per week. |

Your task step is complete only after `submit` returns `"accepted": true`.
\end{promptlisting}

\begin{promptlisting}{Auditing Instructions}
# Auditing task

You are auditing one numeric fact in an SEC XBRL filing (10-K or 10-Q). Your
assignment from `get_task` names the filing, the concept, the reporting period,
and, where needed to identify a single fact, the dimension members and/or the
financial statement.

You must determine two values and submit them with `submit`:

1. **reported_value**: the value of the assigned fact exactly as it is reported
   in the filing's instance document, for the assigned period (and dimensions,
   if any; otherwise the fact without dimensions).

2. **calculated_value**: the value the assigned fact should have for the filing
   to be internally consistent and to follow XBRL reporting conventions,
   determined from the filing's own linkbases and the US-GAAP taxonomy.
   Relevant considerations include:
   - calculation relationships in the filing's calculation linkbase (weighted
     sums of related concepts for the same period and context);
   - dimensional relationships in the filing (an aggregate value and the
     members of an axis presented on a statement);
   - the concept's definition, balance type and sign conventions in the
     US-GAAP taxonomy.
   If you find no inconsistency, the calculated value equals the reported value.

## Tools
All information about the filing comes from the tools of this server. You
cannot access files, the network, or other cases.

| Tool | Use |
|---|---|
| `get_task()` | Your assignment and these instructions. |
| `search_concepts(query)` | Find concepts used in the filing by name or label words. |
| `get_facts(concept_id, period?, dimensions?)` | Reported facts for a concept, with period, dimensions, unit and decimals. |
| `get_concept_metadata(concept_id)` | Labels, balance type, period type, data type and taxonomy definition. |
| `get_calculation_relationships(concept_id)` | Calculation-linkbase parents/children (with weights) of a concept, per statement. |
| `get_dimensional_relationships(concept_id)` | Axes and members that apply to a concept, per statement (definition linkbase). |
| `list_statements(query?)` | Statements and disclosures (presentation roles) in the filing. |
| `get_statement_presentation(statement)` | The presentation hierarchy of one statement. |
| `submit(reported_value, calculated_value)` | Submit your answer. Accepted exactly once. |

Periods are written `YYYY-MM-DD` (instant) or `YYYY-MM-DD to YYYY-MM-DD`
(duration). Concept ids use the `prefix:LocalName` form, e.g.
`us-gaap:AssetsCurrent`.

## Submission format
Both values are plain decimal numbers as strings, in the units of the fact as
reported (no thousands separators, no currency symbols, no scaling), e.g.
`"-1284"`, `"1284"`, `"0.0293"`. Your task is complete only after `submit`
returns `"accepted": true`.
\end{promptlisting}

\begin{promptlisting}{Rubric Generation Prompt}
# Task

Your objective is to identify the most discriminative criteria that distinguish high-quality reports from low-quality ones under a specific evaluation dimension. You will analyze two high-quality Reference Reports, capture subtle quality differences, and formulate atomic, objective, and Ground-Truth-anchored rubrics.

# Input Context

- [Dimension Name]: {dimension}
- [Dimension Description]: {dimension_desc}
- [Report 1]: {report_1}
- [Report 2]: {report_2}

# Categories of Rubrics

You must generate two distinct categories of rubrics:

1. Positive Rubrics: Excellence indicators and fundamental requirements that distinguish superior, highly detailed responses.

2. Negative Rubrics: Critical flaws or active mistakes that definitively degrade the quality of a report. Focus on clear failure modes, not merely the absence of excellence.

# Core Guidelines and Methodologies

You must strictly adhere to the following principles when extracting and generating rubrics.

## 1. Discriminative Power and Methodology

- Consensus Extraction: Identify essential facts, exact data points, or logic steps shared by all reference reports.

- Divergence Resolution: Where reports differ, identify the superior approach that meaningfully separates exceptional quality from mediocre quality.

- Exclude Generic Criteria: Do not generate broad rubrics that apply equally to any standard text.

## 2. Atomic and Verifiable

- Each rubric must contain only one indivisible checkpoint.

- It must be resolvable with a clear Yes/No judgment or require extracting an exact value or string found in the reference reports.

- Never use subjective terms such as "is it detailed?", "is it clear?", or "does it make sense?".

## 3. Avoid Mirror Rubrics

- Never create positive and negative versions of the same criterion.

- For example, do not create "Calculates WACC" as a positive rubric and "Fails to calculate WACC" as a negative rubric.

- Choose only the more discriminative direction.

## 4. Conservative Negative Rubrics

- Negative rubrics should identify active mistakes, not merely missing features.

- Examples include "Conflates revenue with profit when citing Q3 data" or "Cites a future date beyond the report context".

## 5. Quality over Pure Quantity

- Focus on observable, actionable, and highly impactful criteria.

- A highly discriminative and specific rubric is preferable to multiple generic rubrics.

- Aim for exhaustive coverage of the dimension without sacrificing discriminative power.

# Output Format

Output only a valid JSON object with the required structure. Do not include explanatory text before or after the JSON. Do not wrap the JSON in Markdown code fences.
\end{promptlisting}

\section{Evaluation Methods}
\label{sec:sec_evaluation_methods}

\subsection{Evaluation Metrics}
\label{sec:sec_evaluation_metrics}

\paragraph{Market-Based Strategy Returns}
For Trading, Hedging, and Market Insights, we replay the committed
decisions against realized prices. The agent decides before the market
opens on each decision date, and the resulting position is held from
the open of that date to the open of the next trading date (on the last
date of the window, from the open to the close). We use split- and
dividend-adjusted prices, with the adjusted open
\(\tilde o_{i,t}=o_{i,t}\,p_{i,t}/c_{i,t}\) computed from the open
\(o_{i,t}\), close \(c_{i,t}\), and adjusted close \(p_{i,t}\) of asset
\(i\) on date \(t\). The return of asset \(i\) over the holding interval
of date \(t\) is
\[
\rho_{i,t}
=
\frac{\tilde o_{i,t+1}-\tilde o_{i,t}}{\tilde o_{i,t}} .
\]
Given the portfolio weights
\(\mathbf{w}_t^A\) implied by agent \(A\)'s committed decisions, the
strategy return is
\[
r_t^A
=
(\mathbf{w}_t^A)^\top\boldsymbol{\rho}_t .
\]
For Trading, \textsc{Buy}, \textsc{Sell}, and \textsc{Hold} correspond
to exposures \(1\), \(-1\), and \(0\), respectively, and a date without a
committed decision is flat. For Hedging, the pair is fixed at its first
selection; \textsc{Long--Short} assigns weights \((+\frac{1}{2},-\frac{1}{2})\)
and \textsc{Short--Long} \((-\frac{1}{2},+\frac{1}{2})\) to the two legs,
yielding unit gross exposure and zero net exposure. \textsc{Hold} and
a missing decision keep the previous weights, while \textsc{Close} sets
both weights to zero. For Market Insights, each weekly rating is mapped
to an exposure in the asset, \(\omega(\textsc{Strong Buy})=1\),
\(\omega(\textsc{Buy})=\frac{1}{2}\), \(\omega(\textsc{Hold})=0\),
\(\omega(\textsc{Sell})=-\frac{1}{2}\), and
\(\omega(\textsc{Strong Sell})=-1\), which takes effect on the first
trading date after the report date and is held until the next report
takes effect; a missing report is flat, and the evaluation window starts
on the first trading date after the first report. We exclude transaction
costs and other execution frictions.
The Buy\&Hold baseline assigns unit long exposure to the corresponding
asset throughout the same evaluation window.

\paragraph{Cumulative Return (CR)}
Starting from \(W_0=1\), the strategy value evolves as
\[
W_t
=
W_{t-1}(1+r_t^A).
\]
Cumulative return is reported as a percentage:
\[
\mathrm{CR}
=
100\,(W_T-1).
\]

\paragraph{Sharpe Ratio (SR)}
Let \(\bar r\) and \(s_r\) denote the mean and sample standard deviation of
the daily strategy-return sequence. With the daily risk-free rate set
to zero, the annualized Sharpe ratio is
\[
\mathrm{SR}
=
\sqrt{252}\,
\frac{\bar r}{s_r}.
\]

\paragraph{Maximum Drawdown (MDD)}
Let
\[
M_t
=
\max_{0\leq u\leq t} W_u
\]
denote the running peak of the strategy value. Maximum drawdown is
reported as a percentage:
\[
\mathrm{MDD}
=
100
\max_{0\leq t\leq T}
\frac{M_t-W_t}{M_t}.
\]

\paragraph{Auditing Outcomes}
Auditing instances are assigned to one of four mutually exclusive
outcomes through the ordered checks described in
Section~\ref{sec:sec_auditing}. An instance receives a schema error if
the agent fails to produce a valid committed output, an extraction
error if the submitted reported value is incorrect, and a calculation
error if the reported value is correct but the reconstructed value is
incorrect. An instance is correct only if all checks succeed.

Let \(N_A\), \(N_S\), \(N_E\), and \(N_C\) denote the numbers of
correct instances, schema errors, extraction errors, and calculation
errors, respectively. For \(N\) total instances, we report
\[
\begin{aligned}
\mathrm{ACC} &= \frac{N_A}{N},
&
\mathrm{SER} &= \frac{N_S}{N},\\
\mathrm{EER} &= \frac{N_E}{N},
&
\mathrm{CER} &= \frac{N_C}{N}.
\end{aligned}
\]
Because each instance is assigned according to its first failed check,
\[
N_A+N_S+N_E+N_C=N,
\]
and therefore
\[
\mathrm{ACC}
+\mathrm{SER}
+\mathrm{EER}
+\mathrm{CER}
=1.
\]

\subsection{Evaluation Rubrics}
\label{sec:sec_evaluation_rubrics}

Portfolio outcomes capture the realized consequences of the ratings
produced in Market Insights, but do not directly assess the quality of
the supporting reports. We therefore evaluate each report using
instance-specific rubrics covering three dimensions: content accuracy,
evidence fidelity, and reasoning quality. The
evaluation pipeline consists of rubric generation and rubric-based scoring.

\paragraph{Rubric Generation}
For each asset, report week, and dimension, a rubric generator receives
two reference reports for that week and proposes atomic positive and
negative rubric items. Positive items describe specific properties that
a strong report should satisfy, whereas negative items identify active
errors that reduce report quality. The generator is sampled three times
and a final consolidation call removes duplicate, mirrored, subjective,
or non-atomic items. Rubrics are generated separately for three
dimensions:

\begin{enumerate}
    \item \textbf{Content Accuracy}: correctness of dates, metadata,
    and factual claims against the underlying benchmark records.
    
    \item \textbf{Evidence Fidelity}: consistency of quantitative and
    qualitative claims with the prices, news, and filings available for
    the corresponding instance.
    
    \item \textbf{Reasoning Quality}: consistency between evidence,
    investment thesis, rating, risks, and outlook across report
    sections.
\end{enumerate}

For the offline window, the reference reports were written by Claude
Sonnet~4.6 and GPT-5.4 agents, the generator is Claude Sonnet~4
(samplings at temperatures \(0.70\), \(0.75\), and \(0.80\); consolidation
at \(0.2\)), and numeric values in the items are re-anchored to the
benchmark's weekly statistics; items about metrics the environment does
not provide are dropped. For the post-release window, the reference
reports are produced by Claude Sonnet~5 and GPT-5.4 on the
\FAB Market Insights task itself and the generator is Claude Opus~5.5 at
its default temperature, because the earlier generator is no longer
served. The rubrics are generated automatically and are not manually
validated. The generation prompt is provided in
Appendix~\ref{sec:sec_prompts}.

\paragraph{Rubric-Based Evaluation}
We use grok-4.7 as the rubric verifier. For each submitted
report, the verifier evaluates every retained item and returns a binary
judgment. Let \(n_{i,k,d}\) denote the number of rubric items for asset
\(i\), report week \(k\), and dimension \(d\), and let
\(z_{i,k,d,\ell}^{A}\in\{0,1\}\) indicate whether report
\(r_{i,k}^{A}\) receives credit for item \(\ell\). The dimension score
is
\[
q_d(r_{i,k}^{A})
=
\frac{10}{n_{i,k,d}}
\sum_{\ell=1}^{n_{i,k,d}}
z_{i,k,d,\ell}^{A}.
\]
The report score is the equal-weighted average across the three
dimensions, and a missing report scores \(0\):
\[
Q_{i,k}^{A}
=
\frac{1}{3}
\sum_{d=1}^{3}
q_d(r_{i,k}^{A}),
\]
and the asset-level score reported in the main text is
\[
Q_i^A
=
\frac{1}{K}
\sum_{k=1}^{K}
Q_{i,k}^{A}.
\]

\paragraph{Judge Agreement}
The verifier, grok-4.7, comes from a vendor with no backbone in the
study. To check its reliability, a second verifier from another vendor,
gemini-3.8-flash, re-scored all reports of the first run of every
configuration (\(1{,}391\) reports, \(58{,}374\) rubric items). The two
verifiers agree on \(93.0\%\) of item judgments (Cohen's
\(\kappa=0.73\)); report scores correlate at Pearson \(r=0.81\), and the
ranking of the \(30\) configurations by mean score at Spearman
\(\rho=0.96\). The second verifier is more lenient on average (\(+0.6\)).

\section{Analysis Methods for the Main Findings}
\label{app:analysis}

This appendix describes the analyses built on top of the evaluation results in Table~\ref{tab:results}: the normalized workflow scores and architecture ranks used in RQ1 and RQ3 (\S\ref{app:rq1_rq3_scores}), and the behavioral metrics, trajectory statistics, and regression used in RQ2 (\S\ref{app:rq2_metrics}).

\subsection{Workflow Scores and Architecture Ranks}
\label{app:rq1_rq3_scores}

\paragraph{Normalized scores (Figure~\ref{fig:rq1_capability_landscape})}
Each metric in Table~\ref{tab:results} is converted to a percentile rank among the $n=30$ agent--backbone configurations. For metric $m$ (and asset, where applicable), let $r_c$ be the rank of configuration $c$ among the configurations with a valid result (1 = best; ties receive their average rank; MDD is ranked in reverse because lower is better). The normalized score is
\[
s_c = 100\,\frac{n_m - r_c}{n_m - 1},
\]
where $n_m$ is the number of valid configurations; configurations without a valid result receive $s_c=0$. Workflow scores average these normalized scores: over CR, SR, and MDD on each of the four assets for Trading; over CR, SR, and MDD for Hedging; for Market Insights, the mean report-quality score and the mean of CR, SR, and MDD over the four assets, weighted equally; and ACC for Auditing. The overall score is the mean of the four workflow scores, and the workflow floor is their minimum.

\paragraph{Ranking agreement across workflows}
Agreement between workflow rankings is measured by Kendall's $\tau$ between the workflow scores of the 30 configurations, computed for each pair of workflows.

\paragraph{Architecture ranks (Figure~\ref{fig:rq3_architecture_reranking})}
Within each backbone, the five architectures are ranked on each workflow score (1 = best; ties receive their average rank). The plotted value is the mean of these ranks over the six backbones.

\subsection{Behavioral Metrics and Construction of Figure~\ref{fig:rq2_behavioral_traces}}
\label{app:rq2_metrics}

\paragraph{Trajectories}
The offline evaluation window contains $T=61$ trading days (January~2 to March~31, 2026) for Trading and Hedging and $K=13$ report weeks for Market Insights. The decision at step $t$ is encoded as a position or exposure $w_t$: for Trading, $\textsc{Buy}=+1$, $\textsc{Hold}=0$, and $\textsc{Sell}=-1$; for Market Insights, the rating exposure $\omega$ of Appendix~\ref{sec:sec_evaluation_metrics} ($\textsc{Strong Buy}=1$, $\textsc{Buy}=0.5$, $\textsc{Hold}=0$, $\textsc{Sell}=-0.5$, $\textsc{Strong Sell}=-1$); and for Hedging, after the pair is selected, $\textsc{Long-Short}=+1$, $\textsc{Short-Long}=-1$, and flat $=0$, where \textsc{Hold} keeps the previous position and \textsc{Close} returns it to flat. A step without a committed decision follows the rule of the return computation: it is flat in Trading and Market Insights and keeps the previous position in Hedging. Panels~(a) and~(b) draw such a step as \textsc{Hold}, which is equivalent under these rules.

\paragraph{Panel (a): case selection}
Among all backbone--asset combinations in which all five architectures committed a record on every trading day, we compute each architecture's switch rate (the share of consecutive days on which the action changes), average it over architectures, and select the combination closest to the median of this average. We use the first of the three runs. This rule selects DeepSeek-V4-Pro on MSFT and avoids showing a visually extreme example. The top line is the trailing 5-day return of MSFT, $100\,(p_t/p_{t-5}-1)$, computed from adjusted closing prices within the evaluation window. Thin vertical lines mark days on which the action changes.

\paragraph{Panel (b): case selection}
We show the backbone--run--pair group chosen by the largest number of architectures: in the third run under GPT-5.4, ReAct, Codex, Hermes, and OpenClaw all selected MSFT/NVDA. Claude Code selected META/MSFT and is therefore not plotted against this spread. The top line is the log price spread
\[
\log p^{\mathrm{MSFT}}_t-\log p^{\mathrm{NVDA}}_t,
\]
standardized by its mean and standard deviation over the full evaluation window. This ex-post standardization is for display only and was not available to the agents.

\paragraph{Panel (c): metric definitions}
Each metric is computed within a run by pooling all cases (assets) of a configuration, and then averaged over the three runs.
\begin{itemize}
\item \textbf{Reactivity} (Trading, Hedging, Market Insights) is
\[
\mathrm{Reactivity}=\rho\bigl(w_t,r^{\mathrm{past}}_t\bigr)-\rho\bigl(w_t,r^{\mathrm{future}}_t\bigr),
\]
where $\rho$ is the Pearson correlation over all steps of the run. $r^{\mathrm{past}}_t$ is the most recent return visible at the decision: the close-to-close return on the information cut-off date for Trading and Hedging, and the return of the report week for Market Insights. $r^{\mathrm{future}}_t$ is the return over the interval in which the position is held, exactly as in the CR and SR computation. Hedging uses the spread return $r_A-r_B$ of the selected pair in both terms. Positive values indicate decisions that follow past returns rather than anticipate future ones; the metric is undefined when the position never changes within a run.
\item \textbf{Persistence} (Trading, Hedging, Market Insights) is one minus the normalized turnover,
\[
\mathrm{Persistence}=1-\frac{1}{T-1}\sum_{t=2}^{T}\frac{|w_t-w_{t-1}|}{2},
\]
which equals $1$ for a position that never changes and $0$ when every step flips from a full position in one direction to the other.
\item \textbf{Revision} (Market Insights) is the rating revision rate, the share of consecutive pairs of committed reports $(t-1,t)$ whose ratings differ, $\Pr[g_t\neq g_{t-1}]$.
\item \textbf{Protocol completion} (all four workflows) is the share of steps with a valid submission. Incomplete steps are classified by their final status: finished without a submission, a turn, tool-call, or token limit reached, timeout, rejection by the model provider, or termination after such a rejection.
\end{itemize}

\paragraph{Aggregation and uncertainty}
For each metric, the run-averaged values are z-standardized across the five architectures within each backbone and workflow, using the population standard deviation. This removes differences in scale across backbones and workflows, so that the scores reflect differences between architectures. For each architecture and backbone, z-scores are averaged over the applicable workflows, and the point estimate is the mean over the six backbones. The 95\% interval is a percentile interval from a cluster bootstrap that resamples backbones with replacement (10{,}000 resamples). Because only six clusters are resampled, the intervals are approximate and should be read as indicating direction rather than precise uncertainty.

\paragraph{Trading action statistics}
For each Trading trajectory, the \textsc{Hold} share is the fraction of committed actions that are \textsc{Hold}, and the switch rate is the fraction of consecutive committed records whose actions differ. For the action--return correlation, actions are coded as $\textsc{Buy}=+1$, $\textsc{Hold}=0$, and $\textsc{Sell}=-1$, and we compute the Spearman correlation $\rho$ between the action code and (i) the trailing 5-day return and (ii) the next-day return. Each statistic is averaged over an architecture's trajectories across all backbones, assets, and runs; the main text reports the range of these averages across architectures.

\paragraph{Report quality and decision outcome}
For Market Insights, we regress the cumulative return of the committed ratings on the report-quality score using ordinary least squares with asset, architecture, and backbone fixed effects, over all 120 configuration--asset pairs, using the run-averaged score and return. The coefficient on report quality is $\beta=-0.249$ ($p=0.19$). Without these controls, report quality and return are also essentially uncorrelated (Spearman $\rho=0.08$, $p=0.40$).

\end{document}